\documentclass[10pt,twocolumn,letterpaper]{article}

\usepackage{iccv}
\usepackage{times}
\usepackage{graphicx}
\usepackage{amsmath}
\usepackage{amssymb}

\usepackage{booktabs}

\usepackage{cite}
\usepackage{multirow}
\usepackage{verbatim}
\usepackage{mathtools}
\usepackage{xcolor}

\usepackage[title]{appendix}

\usepackage{hyperref}
\hypersetup{pagebackref,breaklinks,letterpaper, colorlinks}

\newcommand{\bZ}{\ensuremath{{\mathbf{Z}}}}

\newcommand{\bA}{\ensuremath{{\mathbf{A}}}}
\newcommand{\bD}{\ensuremath{{\mathbf{D}}}}
\newcommand{\bR}{\ensuremath{{\mathbf{R}}}}

\newcommand{\bU}{\ensuremath{{\mathbf{U}}}}
\newcommand{\bI}{\ensuremath{{\mathbf{I}}}}

\newcommand{\bQ}{\ensuremath{{\mathbf{Q}}}}
\newcommand{\bK}{\ensuremath{{\mathbf{K}}}}
\newcommand{\bV}{\ensuremath{{\mathbf{V}}}}

\newcommand{\bL}{\ensuremath{{\mathbf{L}}}}
\newcommand{\bM}{\ensuremath{{\mathbf{M}}}}
\newcommand{\bN}{\ensuremath{{\mathbf{N}}}}

\newcommand{\bE}{\ensuremath{{\mathbf{E}}}}
\newcommand{\bB}{\ensuremath{{\mathbf{B}}}}

\usepackage[capitalize]{cleveref}
\crefname{section}{Sec.}{Secs.}
\Crefname{section}{Section}{Sections}
\Crefname{table}{Table}{Tables}
\crefname{table}{Tab.}{Tabs.}

\iccvfinalcopy % *** Uncomment this line for the final submission

\def\iccvPaperID{3451} % *** Enter the ICCV Paper ID here

\ificcvfinal\fi

\begin{document}

%%%%%%%%% TITLE
\title{Versatile Depth Estimator Based on Common Relative Depth Estimation and Camera-Specific Relative-to-Metric Depth Conversion}

\author{Jinyoung Jun\\
Korea University\\
{\tt\small jyjun@mcl.korea.ac.kr}
\and
Jae-Han Lee\\
Gauss Labs Inc.\\
{\tt\small jaehan.lee@gausslabs.ai}
\and
Chang-Su Kim\\
Korea University\\
{\tt\small changsukim@korea.ac.kr}
}

\maketitle
\ificcvfinal\thispagestyle{empty}\fi

%%%%%%%%% ABSTRACT
\begin{abstract}
A typical monocular depth estimator is trained for a single camera, so its performance drops severely on images taken with different cameras. To address this issue, we propose a versatile depth estimator (VDE), composed of a common relative depth estimator (CRDE) and multiple relative-to-metric converters (R2MCs). The CRDE extracts relative depth information, and each R2MC converts the relative information to predict metric depths for a specific camera. The proposed VDE can cope with diverse scenes, including both indoor and outdoor scenes, with only a 1.12\% parameter increase per camera. Experimental results demonstrate that VDE supports multiple cameras effectively and efficiently and also achieves state-of-the-art performance in the conventional single-camera scenario.

\end{abstract}
%%%%%%%%% BODY TEXT
\section{Introduction}
\label{sec:introduction}
Monocular depth estimation is a task to regress pixelwise depths from a single image. It provides the 3D layout of a scene, which is useful in applications including autonomous driving \cite{geiger2012we}, pose estimation \cite{ye2011accurate}, and 3D photography \cite{shih20203d}. It is, however, ill-posed since different 3D scenes can be projected onto the same 2D image. Nevertheless, monocular depth estimation is important because only a single camera is available in many applications.

Recently, learning-based algorithms using convolutional neural networks (CNNs) have driven performance improvements in monocular depth estimation \cite{eigen2014depth, laina2016deeper, fu2018deep, hu2019revisiting, lee2019monocular, lee2020multi}. They attempt to overcome the ill-posedness using big training data \cite{silberman2012indoor, geiger2012we}. Moreover, recent depth estimators \cite{bhat2021adabins, yuan2022newcrfs} based on the transformer architecture \cite{dosovitskiy2020image, liu2021swin} provide even better performance than the CNN-based ones.

Since depth labels are affected by the field of view and the range of a depth camera, learning-based depth estimators are often tailored for a specific camera only; they are not `versatile' and should be retrained using a new dataset to be used for a different camera. Also, if they are trained with depth labels from different cameras, their performances are degraded in general. In particular, joint depth estimation of both indoor and outdoor scenes has been considered incompatible for a single network. To overcome this issue, relative depth estimation, which predicts the relative depth order among pixels instead of absolute (or metric) depths, has been studied \cite{zoran2015learning, chen2016single, xian2020structure, lienen2021monocular}. Since the depth order is camera-invariant, relative depth estimators can be trained using heterogeneous depth data from various sources \cite{ranftl2020}. However, they do not yield actual depths required in applications, so they should be fine-tuned for a specific camera to generate metric depths \cite{ranftl2021vision,jun2022depth}.

\begin{figure}[t]
  \centering
%   \fbox{\rule{0pt}{1.6in} \rule{0.9\linewidth}{0pt}}
   \includegraphics[width=\linewidth]{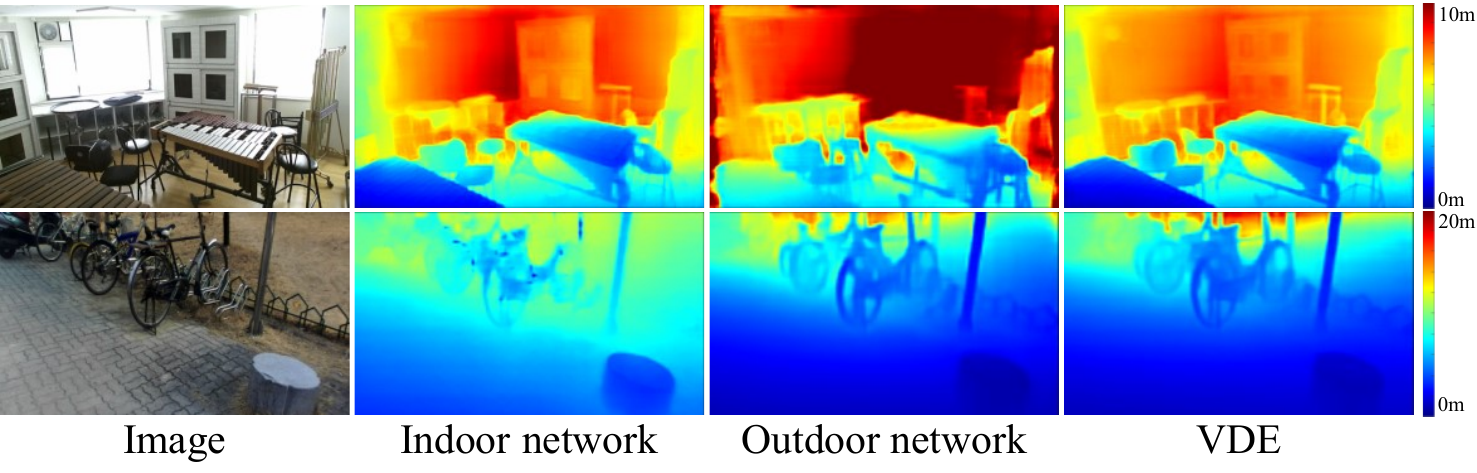}
   \caption{Versatile depth estimation: The indoor network, trained with indoor images only, performs poorly on an outdoor image. Similarly, the outdoor network is less effective for an indoor image. In contrast, the proposed VDE can estimate depths for both indoor and outdoor images reliably in a `versatile' manner.}
   \label{fig:intro}
   \vspace*{-0.25cm}
\end{figure}

In this work, we propose a versatile depth estimator (VDE) to predict depth maps for scenes captured by different cameras, as illustrated in Figure~\ref{fig:intro}. To this end, we design a common relative depth estimator (CRDE) and multiple relative-to-metric converters (R2MCs). First, the CRDE, composed of a transformer encoder and a transformer decoder, extracts relative depth information, which is camera-invariant. It adopts frequency mixing modules (FMMs) to transfer high-frequency components in encoder features to the decoder so that the decoder can predict a detailed relative depth map reliably. Second, each R2MC is employed to convert the relative depth map to the metric one corresponding to a specific camera. By training multiple R2MC modules, the proposed VDE can provide depth maps for multiple cameras. The versatility of VDE is demonstrated by extensive experiments on diverse datasets collected with seven different cameras. Moreover, VDE provides excellent metric depth estimation performance on the NYUv2 dataset \cite{silberman2012indoor}.

This paper has the following contributions:
\begin{itemize}
\itemsep0em
  \item We propose a versatile algorithm, called VDE, for monocular depth estimation, by developing the CRDE and R2MC modules.
  \item VDE yields excellent performance in diverse scenarios using different cameras, with only 1.7M parameters for each R2MC.
  \item VDE provides state-of-the-art performance on NYUv2 in the conventional single-camera scenario, as well as performing efficiently and effectively in the versatile scenarios.
\end{itemize}
%-------------------------------------------------------------------------
\section{Related Work}
\label{sec:related_work}
\subsection{Monocular depth estimation}
In monocular depth estimation, we infer the absolute distance of each scene point in a single image from the camera. Early monocular depth estimation focused on finding rough 3D layouts of scenes based on prior assumptions, such as superpixels \cite{saxena2008make3d}, block world \cite{gupta2010eccv}, or line segments and vanishing points \cite{gupta2010nips}. Such assumptions, however, may be invalid, so unreliable depths may be obtained especially for small objects or regions with ambiguous colors.

CNNs have been adopted for monocular depth estimation successfully. Various attempts have been made to find effective network architecture \cite{eigen2014depth, laina2016deeper, xu2017multi, heo2018monocular, chen2019structure} or loss functions \cite{eigen2015predicting, chen2016single, laina2016deeper, hu2019revisiting, lee2020multi}. Also, transformer \cite{vaswani2017attention} was recently applied to vision tasks \cite{dosovitskiy2020image}, and its variants have been developed for monocular depth estimation \cite{ranftl2021vision, yuan2022newcrfs}. Besides, exploiting inherent information in 3D geometry, including virtual normal \cite{yin2019enforcing}, depth attention volume \cite{huynh2020guiding}, and depth distribution \cite{bhat2021adabins}, has been tried to improve monocular depth estimation. Also, the depth estimation problem has been reformulated as ordinal regression \cite{fu2018deep}, frequency domain analysis \cite{lee2018single}, and planar coefficient estimation \cite{patil2022p3depth}.

\subsection{Relative depth estimation}
Relative depth estimation aims to learn the depth ranks of pixels in an image \cite{zoran2015learning, chen2016single, xian2020structure}. Since its objective is to determine not the absolute distances but the depth order of pixels, heterogeneous data --- \eg disparities from stereoscopic images \cite{wang2019web, xian2018monocular, xian2020structure}, video frames \cite{ranftl2020}, structure-from-motion (SfM) \cite{li2018megadepth, li2019learning}, and human-annotated ordinal labels \cite{chen2016single} --- can be used jointly to train relative depth estimators. Moreover, relative depths are invariant to camera parameters, so images captured by different cameras can be used for training as well. The diversity of training data can lead to better depth estimation.

In relative depth estimation, the scale-invariant loss \cite{eigen2014depth} and its variants \cite{li2018megadepth, li2019learning, wang2019web, ranftl2020} have been used to cope with the scale ambiguity of depth labels. Recently, listwise ranking \cite{lienen2021monocular}, instead of pairwise ranking, and depth normalization \cite{jun2022depth} have been considered.

\subsection{Relative-to-metric depth conversion}
Whereas relative depth information can be readily obtained from absolute depths in a metric depth map, the opposite conversion from relative depths to metric ones is not straightforward. There have been some attempts for this conversion. For instance, relative depths were fitted to metric depths directly using least-squares in \cite{lienen2021monocular, ranftl2020}, and a relative depth estimator was fine-tuned to a metric depth estimator in \cite{ranftl2021vision}. Also, relative and metric depths were jointly learned through depth map decomposition in \cite{jun2022depth}.

However, \cite{lienen2021monocular, ranftl2020} require a large depth-labeled dataset for fitting. Also, \cite{ranftl2021vision,jun2022depth} fine-tune a pre-trained network to estimate metric depth maps for a single camera only. Hence, they are not versatile, \ie, not applicable to images captured by different cameras. In contrast to these methods, the proposed VDE extracts relative depth information using a single CRDE for diverse cameras and then uses a simple R2MC module to convert the relative information to metric depths for a specific camera. Hence, VDE achieves versatile depth estimation efficiently: to support $N$ cameras, only $N$ R2MCs are required in addition to the CRDE. Moreover, by training the CRDE using diverse datasets, VDE yields comparable or even better performance than non-versatile techniques optimized for specific cameras.

\subsection{Transformer and self-attention}
Inspired by the success of the transformer architecture in natural language processing (NLP) \cite{vaswani2017attention, devlin2018bert}, transformers for vision tasks also have been studied extensively. In transformers, self-attention is performed to emphasize essential information and extract informative features. Many transformer-based neural networks \cite{dosovitskiy2020image, liu2021swin, ranftl2021vision, fang2022msg, chen2021regionvit} have been developed to provide even better performances than CNNs in vision tasks. Also, there have been researches for computing attention between different types of features \cite{carion2020end, cheng2021per}. In this paper, we develop a transformer network for monocular depth estimation composed of an encoder and a decoder, as in U-Net \cite{ronneberger2015u}. In U-Net, skip connections can be used to transfer detailed information from the encoder to the decoder. Similarly, we propose FMMs to combine detailed information from the transformer encoder with the features of the transformer decoder effectively.

%-------------------------------------------------------------------------
\begin{figure*}[!t]
  \centering
   \includegraphics[width=\linewidth]{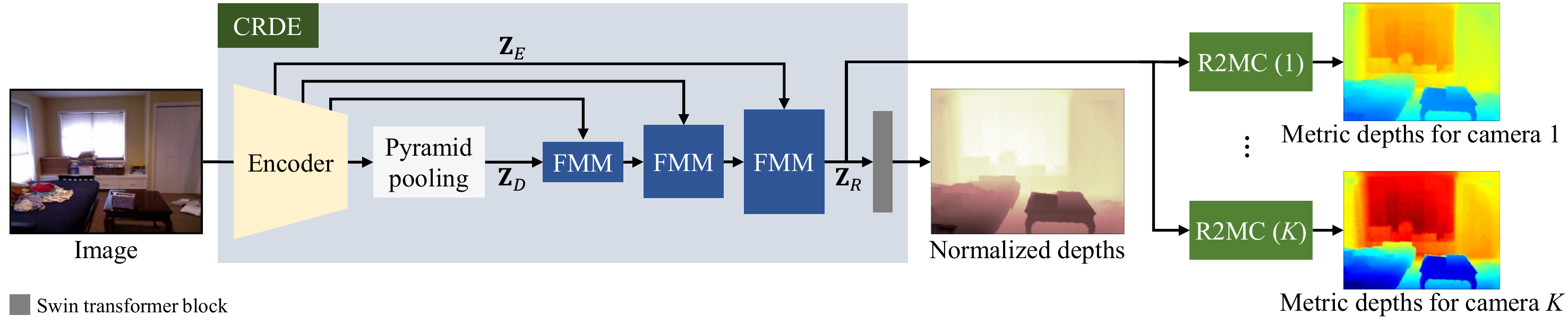}
   \caption{An overview of the proposed VDE: the CRDE extracts the relative depth feature $\bZ_R$, which is processed to yield a normalized depth map. Also, $\bZ_R$ is used to generate a metric depth map by the R2MC for the corresponding camera. }
   \label{fig:overview}
\end{figure*}

\section{Proposed Algorithm}
\label{sec:proposed_algorithm}

In Figure~\ref{fig:overview}, the proposed VDE consists of a single CRDE for extracting relative depth information and multiple R2MCs for converting the information into metric depth maps for different cameras. In the CRDE, the Swin transformer \cite{liu2021swin} is adopted as the encoder, and three FMMs compose the decoder. Multi-scale encoder features are aggregated via pyramid pooling \cite{zhao2017pyramid} and also transferred to the decoder via skip connections. The three FMMs are also based on the Swin transformer, but they mix the decoder feature $\bZ_D$ with the encoder feature $\bZ_E$ adaptively to yield the relative depth feature $\bZ_R$. Then, the CRDE uses $\bZ_R$ to produce a normalized depth map, while each R2MC module uses $\bZ_R$ to yield a metric depth map. We can estimate metric depth maps for scenes captured by $K$ different cameras by training as many R2MCs.

\subsection{Problem formulation}
\label{sec:Versatile_monocular_depth_estimation}

Let $\bI$ be a color image and $\bD$ be its depth map. In monocular depth estimation, an estimator $f$ is trained to minimize the loss
\begin{equation}
    % \textstyle
    \sum_{(\bI, \bD) \in {\cal T}}{\ell(f(\bI), \bD)}
    \label{eq:depth_estimation_formulation}
\end{equation}
where $\cal T$ is a training set of $(\bI, \bD)$ pairs, and $\ell$ is a loss function between an estimated depth map $f(\bI)$ and the ground-truth $\bD$. Versatile monocular depth estimation is more challenging than ordinary one, for it should cope with images captured by diverse cameras with different parameters. Specifically, in versatile depth estimation, ${\cal T}$ in \eqref{eq:depth_estimation_formulation} is a union of multiple datasets, each of which is constructed with a different camera. To solve this problem, we decompose a depth map into camera-invariant and camera-specific components: relative depth information is camera-invariant, whereas depth scale information is camera-specific.

We adopt the normalized depth map as the camera-invariant component, as in \cite{jun2022depth}. Specifically, given a depth map $\bD$, the normalized depth map $\bN$ is obtained by
\begin{equation}
    \textstyle
    \bN = \frac{1}{\sigma}(\bD - \mu \bU)
    \label{eq:depth_normalization}
\end{equation}
where $\mu$ and $\sigma$ are the mean and standard deviation of depths in $\bD$, and $\bU$ is the unit matrix whose elements are all 1. Note that the CRDE estimates a normalized depth map in Figure~\ref{fig:overview}. Thus, the CRDE, denoted by $g(\cdot)$, is trained to minimize the camera-invariant loss
\begin{equation}
    L_{\textrm{CRDE}} = \sum_{(\bI, \bN) \in \bigcup_{n=1}^{K}{\cal T}_n}{\ell(g(\bI), \bN)}
    \label{eq:L_CRDE}
\end{equation}
where ${\cal T}_n$ is the training set for camera $n$. Since the CRDE is common for all cameras, it is trained using all training sets $\bigcup_{n=1}^K{\cal T}_n$.

Given a normalized depth map $\bN$, we can reconstruct the metric depth map $\bM$ via \eqref{eq:depth_normalization} if $\mu$ and $\sigma$ are known. However, they are unknown in practice, and direct conversion is impossible. Hence, we first convert the relative depth feature $\bZ_R$ into the metric depth feature for camera $n$ and then predict a metric depth map through an R2MC. Let $h_n$ denote the R2MC for camera $n$. It is trained to minimize the loss
\begin{equation}
    L_{\textrm{R2MC}}^{(n)} = \sum_{(\bI, \bM) \in {\cal T}_n} {\ell\left(h_n(\bZ_R (\bI)), \bM\right)}
    \label{eq:L_R2MC}
\end{equation}
where $\bZ_R (\bI)$ denotes the relative depth feature extracted from an image $\bI$.

Finally, the overall loss for VDE is given by
\begin{equation}
    L_{\textrm{overall}} =  L_{\textrm{CRDE}} + \sum_{n=1}^{K} L_{\textrm{R2MC}}^{(n)}
\end{equation}
where the first term is camera-invariant, and the second term is camera-specific.

\begin{figure}[!t]
  \centering
  \includegraphics[width=\linewidth]{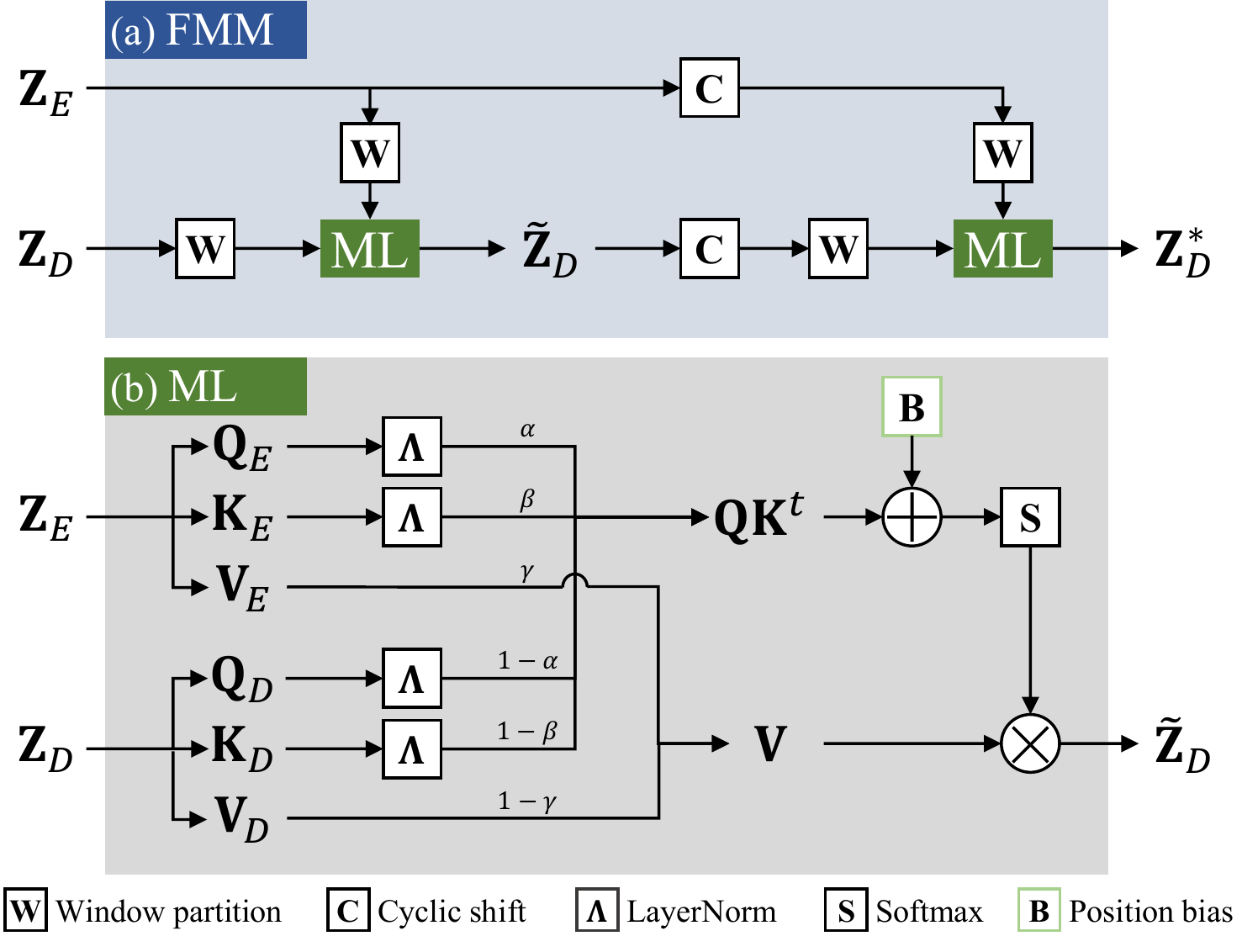}
   \caption{(a) The structure of an FMM and (b) the detailed structure of a mixing layer (ML).}
   \label{fig:FMM_ML}
\end{figure}

\subsection{CRDE}
In monocular depth estimation, the encoder-decoder architecture in U-Net \cite{ronneberger2015u} is widely adopted for its efficacy. The encoder extracts a lower-resolution feature from an image, and then the decoder processes the feature to yield a depth map. At the decoder, interpolation or pixel shuffling is done to match an output resolution. However, the low-resolution feature may lose details, so skip connections are often used to pass details from the encoder to the decoder. In Figure~\ref{fig:overview}, the CRDE also has the U-Net architecture, but it is composed of transformer blocks rather than convolution layers. In order to transfer high-frequency details, we employ three skip connections and compose the decoder with three FMMs.

In a typical depth map, planar regions --- \eg walls, floors, and ceilings --- are dominant. Their depths change gradually and determine the overall 3D layout of a scene. Such depths are low-frequency components. In contrast, the depths of small objects or edge regions are mainly high-frequency components. Attempts have been made to reconstruct either low-frequency or high-frequency components reliably. For instance, in \cite{heo2018monocular}, the whole strip masking was proposed to detect low-frequency planar regions. On the other hand, to recover high-frequency details, gradient losses \cite{hu2019revisiting, lee2020multi} and contour information \cite{ramamonjisoa2019sharpnet} have been exploited. In this paper, we develop FMMs to combine high-frequency details from the transformer encoder with reliable low-frequency information in the transformer decoder effectively.

Figure~\ref{fig:FMM_ML} shows the structure of an FMM, which combines a decoder feature $\bZ_D \in \mathbb{R}^{C_D \times \frac{H}{2} \times \frac{W}{2}}$ with an encoder feature $\bZ_E \in \mathbb{R}^{C_E \times H \times W}$. Note that the spatial resolution of $\bZ_D$ is half of that of $\bZ_E$. Hence, $\bZ_D$ contains relatively low-frequency components. The FMM mixes $\bZ_D$ with high-frequency details in $\bZ_E$ to yield a refined, high-resolution feature $\bZ^*_D \in \mathbb{R}^{C^*_D \times H \times W}$, given by
\begin{equation}
\begin{array}{l}
\bZ^*_D = \textrm{FMM}(\bZ_E, \bZ_D),\\
\end{array}
\label{eq:FMM}
\end{equation}
through an attention mechanism. An FMM is based on the Swin transformer block, so it contains two mixing layers in Figure~\ref{fig:FMM_ML}(a). The first mixing layer combines $\bZ_E$ and $\bZ_D$ to yield $\widetilde{\bZ}_D$, and then, after the cyclic shift \cite{liu2021swin}, the second mixing layer processes $\bZ_E$ and $\widetilde{\bZ}_D$ to generate $\bZ^*_D$.

\vspace*{0.2cm}

\noindent
\textbf{Self-attention:}
Let us briefly summarize the self-attention mechanism \cite{vaswani2017attention,dosovitskiy2020image} in column vector notations, which are necessary to describe FMMs clearly. In a transformer block, an input feature map $\bZ$ is spatially partitioned  into $N$ tokens and expressed as
\begin{equation}
\bZ = \begin{bmatrix} z_1^t \\ \vdots \\ z_{N}^t \end{bmatrix} \in \mathbb{R}^{N \times C}
\end{equation}
where each token $z_n$, $1\leq n\leq N$, is a $C$-dimensional column vector. Then, $\bZ$ is transformed into the query, key, and value matrices by
\begin{align}
\bQ &= [q_1, \cdots, q_{N}]^t = \bZ\bU^t_Q  \\
\bK &= [k_1, \cdots, k_{N}]^t = \bZ\bU^t_K \label{eq:qkv_definition} \\
\bV &= [v_1, \cdots, v_{N}]^t = \bZ\bU^t_V
\end{align}
where $\bU_Q, \bU_K, \bU_V \in \mathbb{R}^{P \times C}$ are projection matrices. By matching queries with keys, the attention matrix $\bA$ is determined as
\begin{equation}
\bA = \textrm{softmax}\left(\frac{\bQ\bK^t}{\sqrt{N}} + \mathbf{B}\right)
    \label{eq:attention_matrix}
\end{equation}
where $\mathbf{B}$ is a position bias \cite{liu2021swin}. Finally, the self-attended output is given by
\begin{equation}
\bZ^* = \bA \bV.
    \label{eq:self_attention}
\end{equation}

\vspace*{0.2cm}

% \begin{figure}[!t]
%   \centering
%   \includegraphics[width=\linewidth]{./figure/FMM_layers.eps}
%   \caption{(a) The structure of an FMM and (b) the detailed structure of a mixing layer (ML).}
%   \label{fig:FMM_layers}
% \end{figure}

\noindent
\textbf{FMM:}
The encoder feature $\bZ_E$ and the decoder feature $\bZ_D$ are input to a mixing layer in FMM, as shown in Figure~\ref{fig:FMM_ML}(b). They are reshaped into matrices of $N$ tokens, respectively;
\begin{equation}
\bZ_E = \begin{bmatrix} z_{E_1}^t \\ \vdots \\ z_{E_N}^t \end{bmatrix} \in \mathbb{R}^{N \times C_E}, \,\,
\bZ_D = \begin{bmatrix} z_{D_1}^t \\ \vdots \\ z_{D_N}^t \end{bmatrix} \in \mathbb{R}^{N \times C_D'}.
\label{eq:ZeZd}
\end{equation}
Note that two kinds of query matrices can be considered,
\begin{align}
\bQ_E &= \bZ_E \bU_{Q_E}^t \\
\bQ_D &= \bZ_D \bU_{Q_D}^t
\end{align}
where $\bU_{Q_E} \in \mathbb{R}^{P \times C_E}$ and $\bU_{Q_D} \in \mathbb{R}^{P \times C_D'}$. Similarly, we have two key matrices $\bK_E$ and $\bK_D$ and two value matrices $\bV_E$ and $\bV_D$. There are many possibilities for combining these query, value, and key matrices to yield the attended output $\bZ^*$ in \eqref{eq:self_attention}. For example, in \cite{yuan2022newcrfs}, Yuan \etal obtained queries and keys from the encoder feature but values from the decoder feature. In other words, they set $\bQ = \bQ_E$ and $\bK=\bK_E$ in \eqref{eq:attention_matrix} and $\bV=\bV_D$ in \eqref{eq:self_attention} to yield $\bZ^*$.

Instead of the manual choice of query, key, and value matrices, we attempt to learn an optimal way to mix low-frequency base information in $\bZ_D$ with high-frequency details in $\bZ_E$. More specifically, as shown in Figure~\ref{fig:FMM_ML}(b), we determine the query and key matrices by
\begin{align}
\bQ &= \alpha \cdot \Lambda(\bQ_E) + (1 - \alpha) \cdot \Lambda(\bQ_D) \label{eq:Qmix} \\
\bK &= \beta \cdot \Lambda(\bK_E) + (1 - \beta) \cdot \Lambda(\bK_D) \label{eq:Kmix}
\end{align}
where $\alpha$ and $\beta$ are learnable mixing coefficients, and $\Lambda(\cdot)$ denotes the layer normalization \cite{ba2016layer}. Note that elements in $\bZ_E$ may be in a different scale from those in $\bZ_D$ does. For example, if $\bZ_D$ consists of much larger elements than $\bZ_E$ do, the softmax operation in \eqref{eq:attention_matrix} may be dominated by $\bZ_D$ regardless of the relative importance of $\bZ_E$ and $\bZ_D$. To alleviate this scale issue, we adopt the layer normalization in \eqref{eq:Qmix} and \eqref{eq:Kmix}. Also, note that $\bQ \bK^t$ in \eqref{eq:attention_matrix} is composed of four terms: $\bQ_E \bK_E^t$, $\bQ_E \bK_D^t$, $\bQ_D \bK_E^t$, and $\bQ_D \bK_D^t$. Because of $\bQ_E \bK_D^t$ and $\bQ_D \bK_E^t$, the cross-attention between the encoder feature and the decoder feature is performed. On the other hand, $\bQ_E \bK_E^t$ and $\bQ_D \bK_D^t$ are self-attention terms.

We also mix the encoder and decoder features to obtain the value matrix given by
\begin{equation}
\bV = \gamma \cdot \bV_E + (1 - \gamma) \cdot \bV_D \label{eq:Vmix}
\end{equation}
where $\gamma$ is another learnable coefficient. It was observed that normalization is not necessary for $\bV_E$ and $\bV_D$.

By plugging \eqref{eq:Qmix}, \eqref{eq:Kmix}, \eqref{eq:Vmix} into \eqref{eq:attention_matrix} and \eqref{eq:self_attention}, we obtain the refined decoder feature: $\widetilde{\bZ}_D$ in the first mixing layer or $\bZ_D^*$ in the second mixing layer. The output $\bZ_D^*$ of an FMM is used as input to the next FMM. As shown in Figure~\ref{fig:overview}, the output of the last FMM becomes the relative depth feature $\bZ_R$, which is used to estimate a normalized depth map and also converted to a metric depth map by each R2MC module.

\begin{figure}[!t]
  \centering
  \includegraphics[width=\linewidth]{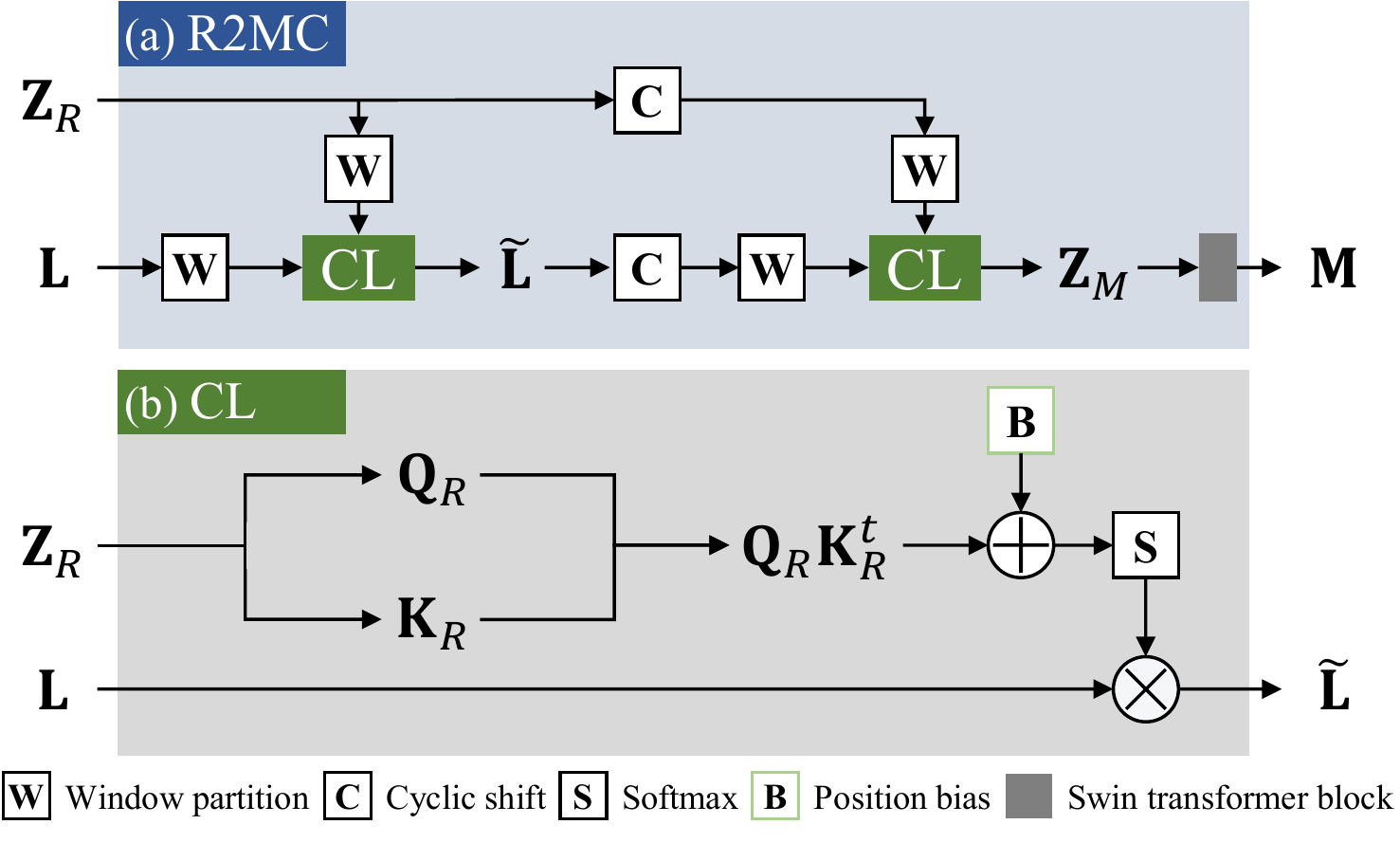}
   \caption{(a) The structure of an R2MC and (b) the detailed structure of a conversion layer (CL).}
   \label{fig:R2MC_CL}
\end{figure}

\subsection{R2MC}

For VDE to support $K$ cameras, we should implement and train $K$ R2MCs. In an R2MC for a specific camera, we first convert the relative depth feature $\bZ_R$ from the CRDE into the metric depth feature $\bZ_M$ optimized for that camera.

Similar to an FMM, an R2MC contains two conversion layers, as shown in Figure \ref{fig:R2MC_CL}(a). The first conversion layer takes the relative depth feature $\bZ_R$ and a learnable matrix $\bL$ as input to generate the output
\begin{equation}
\widetilde{\bL} = \textrm{softmax}\left(\frac{\bQ_R\bK_R^t}{\sqrt{N}} + \bB \right)\bL
\label{eq:R2MC}
\end{equation}
where $\bQ_R$ and $\bK_R$ are the query and key matrices obtained from $\bZ_R$ through the self-attention in Figure~\ref{fig:R2MC_CL}(b). However, instead of the value matrix $\bV_R$, a learnable matrix $\bL$ is used in \eqref{eq:R2MC}. Next, the second conversion layer takes $\bZ_R$ and $\widetilde{\bL}$ to provide the metric depth feature $\bZ_M$. Finally, an ordinary Swin transformer block processes $\bZ_M$ to yield a metric depth map $\bM$.

By adopting a learnable matrix $\bL$ in \eqref{eq:R2MC}, each R2MC learns how to convert the relative depth feature $\bZ_R$ into the camera-specific metric depth feature $\bZ_M$. Thus, $\bL$ plays the role of a camera-specific mapping function from the relative depth space to the metric one. Each R2MC requires only 1.7M parameters, so the proposed VDE can be implemented at the cost of only a moderate increase in complexity.

\subsection{Loss function}
We use the scale-invariant logarithmic loss function $\ell$ for both $L_{\textrm{CRDE}}$ in \eqref{eq:L_CRDE} and $L_{\textrm{R2MC}}$ in \eqref{eq:L_R2MC}. Let $d_i$ and $\hat{d}_i$ denote the $i$th depths in the ground-truth depth map $\bD$ and a predicted depth map $\hat{\bD}$, respectively. Then, the loss function is defined as
\begin{equation}
\ell(\hat{\bD}, \bD)=\alpha \sqrt{\frac{1}{|\bD|}\sum_{i}{{e_i}^2}-\frac{\lambda}{|\bD|^2}(\sum_{i}{e_i})^2}
\label{eq:SILog}
\end{equation}
where $e_i=\log \hat{d}_i - \log d_i$, and $|\bD|$ denotes the number of valid pixels in $\bD$. As in \cite{bhat2021adabins, yuan2022newcrfs}, we set $\alpha=10$ and $\lambda=0.85$.
%-------------------------------------------------------------------------

\begin{table}[t!]
    \scriptsize
    \setlength{\tabcolsep}{3.4pt}
    \caption{Scene types and cameras for the five datasets.}
    \vspace{0.1cm}
    \centering
    % \begin{tabular}{c@{\hskip 1.5em}c@{\hskip 1.5em}c@{\hskip 1.5em}c@{\hskip 1.5em}c}
    \begin{tabular}{ccccc}
    \toprule
    NYUv2 & DIML & DIODE & ScanNet & SUN RGB-D \\
    \midrule
    Indoor & Indoor, Outdoor & Indoor, Outdoor & Indoor & Indoor\\
    \midrule
    \multirow{2}{*}{Kinect v1} & Kinect v2, & \multirow{2}{*}{\shortstack{Laser \\ scanner}} & \multirow{2}{*}{\shortstack{Structure \\ sensor}} & Kinect v1, Kinect v2\\
    & ZED stereo & & & RealSense, Xtion\\
    \bottomrule
    \end{tabular}
    \label{tb:cameras}
\end{table}

\section{Experimental Results}
\label{sec:experimental_results}
\subsection{Datasets}

To assess the versatile depth estimation performance and reliability of the proposed VDE in diverse scenarios, we use five datasets: NYUv2 \cite{silberman2012indoor}, DIML \cite{kim2018deep}, DIODE \cite{vasiljevic2019diode}, ScanNet \cite{dai2017scannet}, and SUN RGB-D \cite{song2015sun}. They are divided into 10 sub-data\-sets according to the scene types and the cameras. Specifically, DIML and DIODE contain indoor and outdoor scenes, so they are, respectively, partitioned into two sub-datasets according to the scene types. Also, SUN RGB-D is divided into four sub-datasets depending on the cameras for the data acquisition. Table~\ref{tb:cameras} summarizes the five datasets. More detailed information on the datasets is provided in the appendix (Section B).

\subsection{Evaluation metrics}
\noindent
\textbf{Metric depth assessment:}
We adopt the top four metrics in Table~\ref{tb:eval_metric} and follow the protocol in \cite{lee2019big, bhat2021adabins, jun2022depth, yuan2022newcrfs}, in which only the regions with valid ground-truth depths are taken into account in the evaluation.

\vspace*{0.1cm}
\noindent
\textbf{Relative depth assessment:}
We use the bottom two metrics in Table~\ref{tb:eval_metric}. First, `Relative $\delta_1$' measures $\delta_1$ between a calibrated relative depth map $\Tilde{\bR}$ and the ground-truth metric depth map $\bD$. As in \cite{ranftl2020, ranftl2021vision}, for a predicted relative depth map $\hat{\bR}$, the calibrated depth map is given by $\Tilde{\bR} = m\hat{\bR} + b\mathbf{1}$, where $\mathbf{1}$ is the depth map consisting of all $1$'s.  For each image, the scaling parameter $m$ and the shift parameter $b$ are adjusted to minimize the least square error between $\Tilde{\bR}$ and $\bD$. Second, Kendall's $\tau$ \cite{kendall1938new} quantifies the confidence of the depth order among pixels by considering the number of concordant and discordant pixel pairs in $\hat{\bD}$ and $\bD$. A higher Kendall's $\tau$ indicates a better result.

\begin{table}[!t]
    \centering
    \small
    \renewcommand{\arraystretch}{1.3}
    \setlength{\tabcolsep}{4pt}
    \caption
    {
        Evaluation metrics for depth maps. The top four are for metric depths, while the bottom two are for relative depths. Here, $|\bD|$ denotes the number of valid pixels in a depth map $\bD$, $d_i$ is the $i$th valid depth in $\bD$, and $\hat{d}_i$ is an estimate of $d_i$. Also, $\operatorname{sgn}(\cdot)$ is the sign function returning 1 for a positive value and 0 otherwise.
    }
    \vspace{0.1cm}
    \begin{tabular}{l l}
    \toprule
    RMSE & $\frac{1}{|\bD|}\big(\sum_{i}(\hat{d}_i-d_i)^2\big)^{0.5}$\\
    \midrule
    REL & $ \frac{1}{|\bD|}\sum_{i} \vert \hat{d}_i-d_i \vert /d_i$\\
    \midrule
    $\log$$10$ & $ \frac{1}{|\bD|}\sum_{i} \vert \log_{10}(\hat{d}_i)-\log_{10}(d_i) \vert$\\
    \midrule
    $\delta_k$, $k=1,2,3$ & \% of $d_i$ that satisfies
    $\max\!\left\{\frac{\hat{d}_i}{d_i},\frac{d_i}{\hat{d}_i}\right\}<1.25^k$\\
    \midrule
    Relative $\delta_1$ & $\delta_1$ between $\Tilde{\bR} = m\hat{\bR} + b\mathbf{1}$ and $\bD$ \\
    \midrule
    % Kendall's $\tau$ & $\frac{\mathcal{C}(\hat{\bD}, \bD) - \mathcal{D}(\hat{\bD}, \bD)}{\binom{|\bD|}{2}}$\\
    Kendall's $\tau$ & $\frac{1}{\binom{|\bD|}{2}}\sum_{i}\sum_{j}\mathrm{sgn}\big((\hat{d}_i - \hat{d}_j)(d_i - d_j)\big)$\\
    \bottomrule
    \end{tabular}
    \label{tb:eval_metric}
\end{table}

\subsection{Implementation details}
\noindent
\textbf{Network architecture:}
In Figure~\ref{fig:overview}, as the encoder of VDE, we adopt Swin-Base \cite{liu2021swin}, which takes an RGB image of spatial resolution $H \times W$ to generate multi-scale features. Then, the pyramid pooling \cite{zhao2017pyramid} combines those features to produce the decoder feature $\bZ_D$ of resolution $\frac{H}{32} \times \frac{W}{32}$. The decoder consists of three FMMs, each of which doubles the spatial resolution. The output of the last FMM is the relative depth feature $\bZ_R$, which is fed to a Swin transformer block to yield a normalized depth map. Also, using $\bZ_R$, each R2MC estimates the metric depth map captured by the corresponding camera. The resolution of the relative depth map and the metric depth maps is $\frac{H}{4} \times \frac{W}{4}$, so we interpolate them bilinearly to the original $H \times W$ resolution.

\vspace*{0.1cm}
\noindent
\textbf{Training:}
VDE is trained in two steps. First, we train only CRDE to estimate normalized depth maps, after initializing the encoder with a pre-trained Swin-Base on ImageNet \cite{liu2021swin}. Second, we train the complete network to produce metric depth maps for different cameras. In both steps, the Adam optimizer \cite{kingma2014adam} is used with a batch size of 4 and a weight decay of $10^{-2}$. The initial learning rate is set to $2 \times 10^{-5}$, and it decreases linearly to $1 \times 10^{-6}$. Normalized depth losses are computed by \eqref{eq:L_CRDE}, while, in the second step, metric depth losses are evaluated separately for different cameras via \eqref{eq:L_R2MC}. The elements of the learnable matrix $\bL$ in each R2MC are initialized to $2 \times 10^{-2}$, and $\alpha, \beta$, and $\gamma$ in each FMM are initialized to 0.5.

\begin{figure}[!t]
  \centering
  \includegraphics[width=\linewidth]{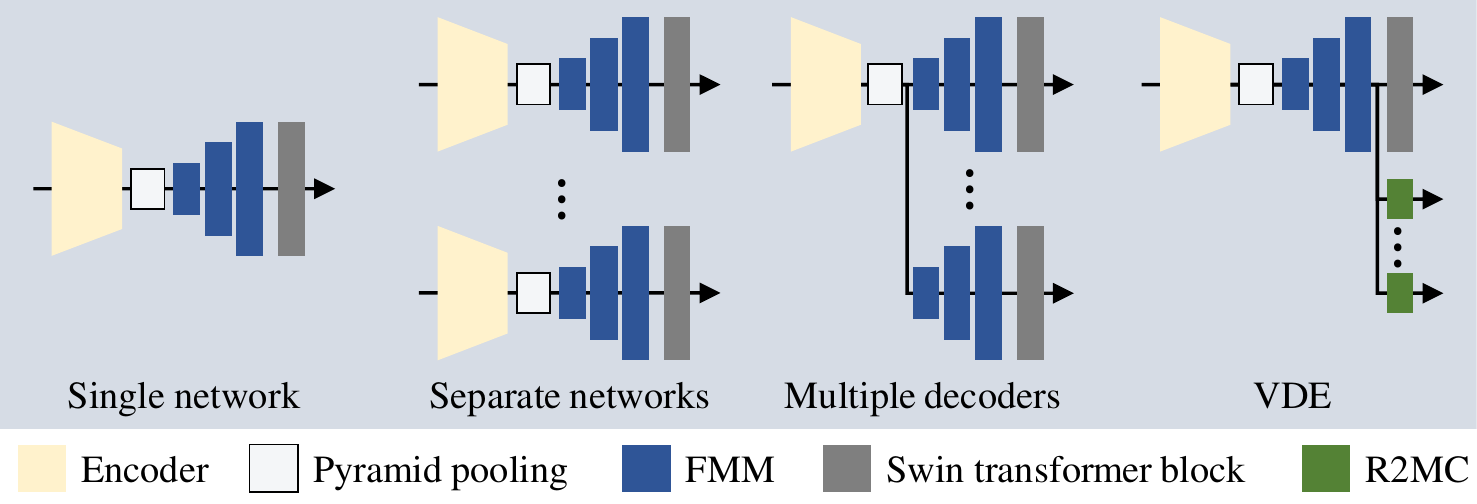}
   \caption{Four network structures for versatile depth estimation.}
   \label{fig:versatile_architecture}
   \vspace*{-0.2cm}
\end{figure}

\subsection{Versatile depth estimation}\label{Versatile_depth_estimation}
We compare versatile depth estimation results of the proposed VDE and three alternative settings. Figure \ref{fig:versatile_architecture} illustrates the network structures of these settings. `Single network' optimizes one CRDE using all 10 sub-datasets, while `Separate networks' optimize 10 CRDEs separately for the 10 sub-datasets. In `Multiple decoders,' a shared encoder is trained jointly with 10 decoders, each of which consists of three FMMs and a Swin transformer block. Finally, VDE is trained using a shared CRDE and 10 separate R2MCs.

Since each sub-dataset has a different size, data imbalance can be incurred. Hence, we sample 1,000 images from each sub-dataset, except for the RealSense sub-dataset of SUN RGB-D, for which we use all 587 images available. VDE is trained for 64 epochs, while `Single network,' `Separate networks,' and `Multiple decoders' are trained for 64, 640, and 64 epochs, respectively. Therefore, the four settings are trained for similar numbers of iterations.

\begin{figure}[t]
    \centering
    \includegraphics[width=\linewidth]{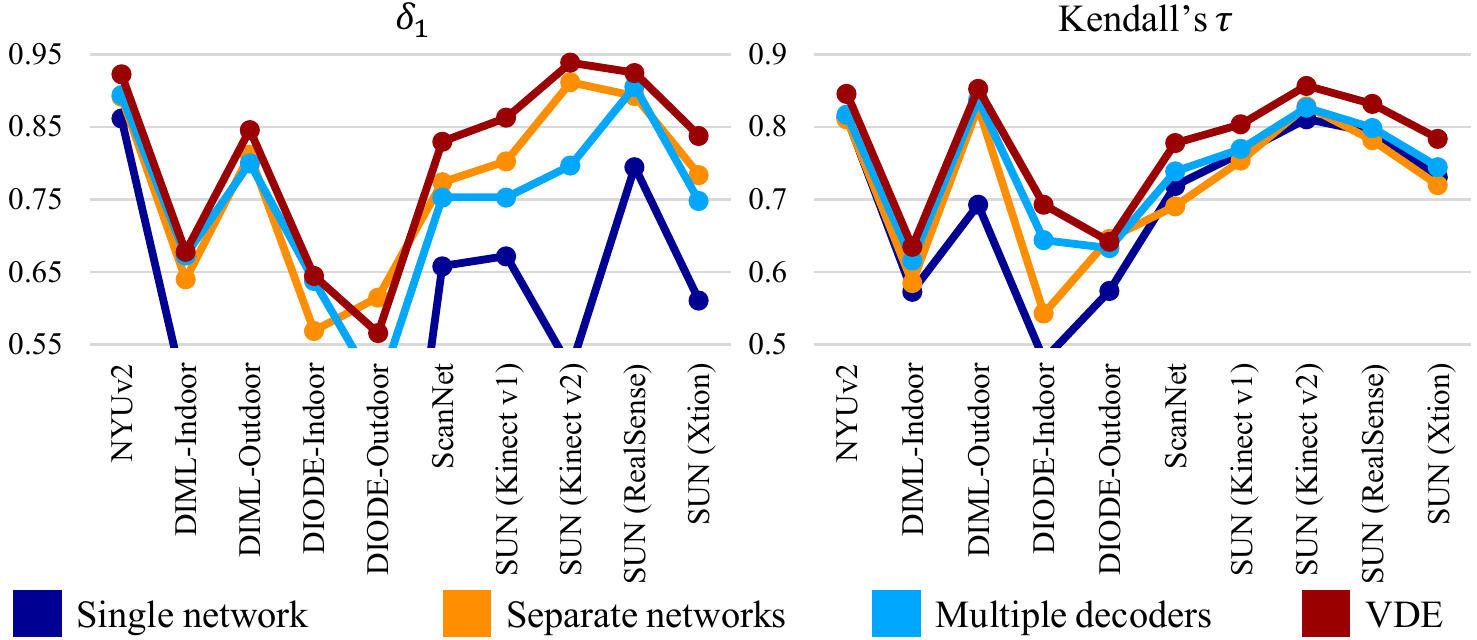}
    \caption{Comparison of $\delta_1$ and Kendall's $\tau$ results of the four settings on the 10 sub-datasets. }
    \label{fig:versatile_performance_graph}
\end{figure}

\begin{table}[!t]
    \scriptsize
    \setlength{\tabcolsep}{4.6pt}
    \caption{Comparison of versatile depth estimation results of `Separate networks,' `Multiple decoders,' and VDE. For each setting, the required number of parameters is specified. Lower RMSE and REL indicate better results, whereas higher $\delta_1$ and $\tau$ are better. In each test, the best result is \textbf{boldfaced}.}
    \vspace{0.1cm}
    \centering
    \begin{tabular}{c|l|ccccc}
    \toprule
    Setting & \multicolumn{1}{c|}{Dataset} & RMSE$(\downarrow)$ & REL$(\downarrow)$ & $\delta_1(\uparrow)$ & $\tau(\uparrow)$\\
    \midrule
    \multirow{10}{*}{\shortstack{Separate\\networks\\(149.8M \!\! $\times$ \!\! 10)}} & NYUv2 & 0.376 & 0.111 & 0.892 & 0.811\\
    & DIML-Indoor & 0.693 & 0.244 & 0.640 & 0.585\\
    & DIML-Outdoor & \textbf{1.240} & \textbf{0.142} & 0.812 & 0.828\\
    & DIODE-Indoor & 1.352 & 0.255 & 0.569 & 0.543\\
    & DIODE-Outdoor & \textbf{5.880} & \textbf{0.302} & \textbf{0.615} & \textbf{0.646}\\
    & ScanNet & 0.347 & 0.169 & 0.774 & 0.691\\
    & SUN (Kinect v1) & 0.324 & 0.157 & 0.803 & 0.754\\
    & SUN (Kinect v2) & 0.257 & 0.099 & 0.912 & 0.829\\
    & SUN (RealsSense) & 0.245 & 0.104 & 0.893 & 0.782\\
    & SUN (Xtion) & 0.405 & 0.416 & 0.784 & 0.720\\
    \midrule
    \multirow{10}{*}{\shortstack{Multiple\\decoders\\(571.3M)}} & NYUv2 & 0.380 & 0.109 & 0.894 & 0.817\\
    & DIML-Indoor & 0.662 & 0.235 & 0.673 & 0.616\\
    & DIML-Outdoor & 1.356 & 0.148 & 0.800 & 0.838\\
    & DIODE-Indoor & 1.180 & \textbf{0.217} & 0.638 & 0.644\\
    & DIODE-Outdoor & 6.973 & 0.351 & 0.485 & 0.633\\
    & ScanNet & 0.340 & 0.185 & 0.753 & 0.739\\
    & SUN (Kinect v1) & 0.349 & 0.184 & 0.753 & 0.770\\
    & SUN (Kinect v2) & 0.330 & 0.168 & 0.797 & 0.827\\
    & SUN (RealsSense) & 0.223 & 0.110 & 0.906 & 0.799\\
    & SUN (Xtion) & 0.417 & 0.467 & 0.748 & 0.745\\
    \midrule
    \multirow{10}{*}{\shortstack{VDE\\(167.3M)}} & NYUv2 & \textbf{0.335} & \textbf{0.093} & \textbf{0.925} & \textbf{0.848}\\
    & DIML-Indoor & \textbf{0.653} & \textbf{0.228} & \textbf{0.678} & \textbf{0.635}\\
    & DIML-Outdoor & 1.245 & 0.146 & \textbf{0.846} & \textbf{0.853}\\
    & DIODE-Indoor & \textbf{1.175} & 0.222 & \textbf{0.645} & \textbf{0.693}\\
    & DIODE-Outdoor & 6.141 & 0.334 & 0.566 & 0.642\\
    & ScanNet & \textbf{0.295} & \textbf{0.139} & \textbf{0.830} & \textbf{0.778}\\
    & SUN (Kinect v1) & \textbf{0.287} & \textbf{0.125} & \textbf{0.863} & \textbf{0.804}\\
    & SUN (Kinect v2) & \textbf{0.231} & \textbf{0.088} & \textbf{0.939} & \textbf{0.857}\\
    & SUN (RealsSense) & \textbf{0.215} & \textbf{0.098} & \textbf{0.925} & \textbf{0.832}\\
    & SUN (Xtion) & \textbf{0.363} & \textbf{0.411} & \textbf{0.838} & \textbf{0.784}\\
    \bottomrule
    \end{tabular}
    \label{tb:performance_versatile}
    \vspace*{-0.1cm}
\end{table}

\begin{figure}[!t]
  \centering
%   \fbox{\rule{0pt}{2in} \rule{0.9\linewidth}{0pt}}
  \includegraphics[width=\linewidth]{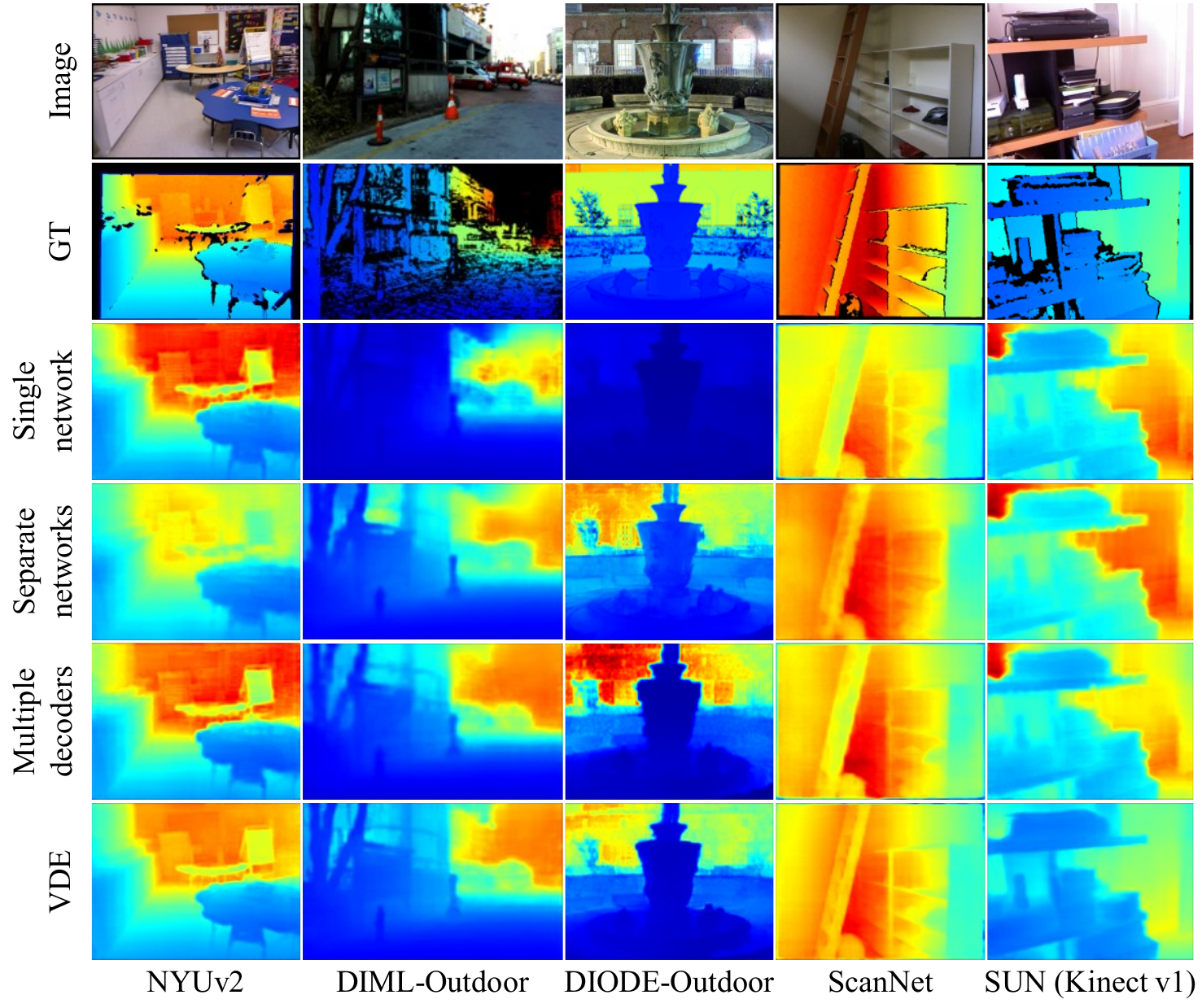}
   \vspace*{-0.5cm}
   \caption{Examples of versatile depth estimation results.}
   \label{fig:qualitative_versatile}
   \vspace*{-0.2cm}
\end{figure}

Figure~\ref{fig:versatile_performance_graph} compares the four settings in terms of $\delta_1$ and Kendall's $\tau$. Also, Table~\ref{tb:performance_versatile} provides detailed comparisons of `Separate networks,' `Multiple decoders,' and the proposed VDE on the 10 sub-datasets. The worst-performing `Single network' is excluded from Table~\ref{tb:performance_versatile}.
The following observations can be made from Figure~\ref{fig:versatile_performance_graph} and Table~\ref{tb:performance_versatile}:
\begin{itemize}
\itemsep0mm
\item By using only 167.3M parameters, VDE generally outperforms `Separate networks' and `Multiple decoders' using 1,498M and 571.3M parameters, respectively. It is worth pointing out that VDE requires 1.7M parameters only to add an R2MC for a new camera.
    \vspace*{-0.05cm}
\item Even when compared with `Separate networks' optimized for each camera, VDE performs better in most tests. Note that VDE is more effective for indoor images than for outdoor ones (DIML-Out\-door and DIODE-Outdoor) because the ratio of indoor to outdoor training images is about 8 to 2 in this experiment. However, even on the outdoor images, VDE yields comparable results to `Separate networks.'
    \vspace*{-0.05cm}
\item On the other hand, on the indoor images, VDE meaningfully outperforms `Separate networks.' For example, in Kendall's $\tau$ on DIODE-Indoor, VDE surpasses `Separate networks' by a large margin of 0.150.
    \vspace*{-0.05cm}
\item Furthermore, VDE performs better than `Multiple decoders' because it trains more parts of the network jointly using heterogenous data with the common goal of estimating normalized depth maps accurately.
\vspace*{-0.05cm}
\item Overall, VDE yields reliable depth estimation results on both indoor and outdoor images. In general, outdoor images have different depth scales from indoor ones. This is a reason why `Single network' performs poorly in Figure~\ref{fig:versatile_performance_graph}. In contrast, VDE overcomes the scale differences between indoor and outdoor scenes by adopting an R2MC for each camera. This is beneficial since a practical algorithm should be capable of handling diverse scenes and cameras.
\end{itemize}
Figure~\ref{fig:qualitative_versatile} shows examples of estimated depth maps. We see that VDE yields more reliable depth maps for diverse images than the alternative settings do.

More experimental results, including cross-dataset evaluation and tests on heterogeneous datasets of different sizes, are in the appendix (Section~C).

\subsection{Ordinary depth estimation}
Although the primary focus of VDE is versatile depth estimation, it achieves state-of-the-art performance in ordinary depth estimation as well. Table \ref{tb:performance_NYU} assesses the metric depth performance of VDE on the NYUv2 dataset. Earlier algorithms \cite{eigen2014depth, wang2015towards, laina2016deeper, fu2018deep, hu2019revisiting, chen2019structure, ramamonjisoa2019sharpnet, yin2019enforcing, lee2020multi} perform evaluation by filling in missing ground-truth depths using a colorization scheme \cite{levin2004colorization}. We distinguish those from recent ones \cite{lee2019big, bhat2021adabins, yang2021transformer, ranftl2021vision, jun2022depth, patil2022p3depth, yuan2022newcrfs} that use valid depths only.

\begin{figure}
  \centering
%   \fbox{\rule{0pt}{3in} \rule{0.9\linewidth}{0pt}}
  \includegraphics[width=\linewidth]{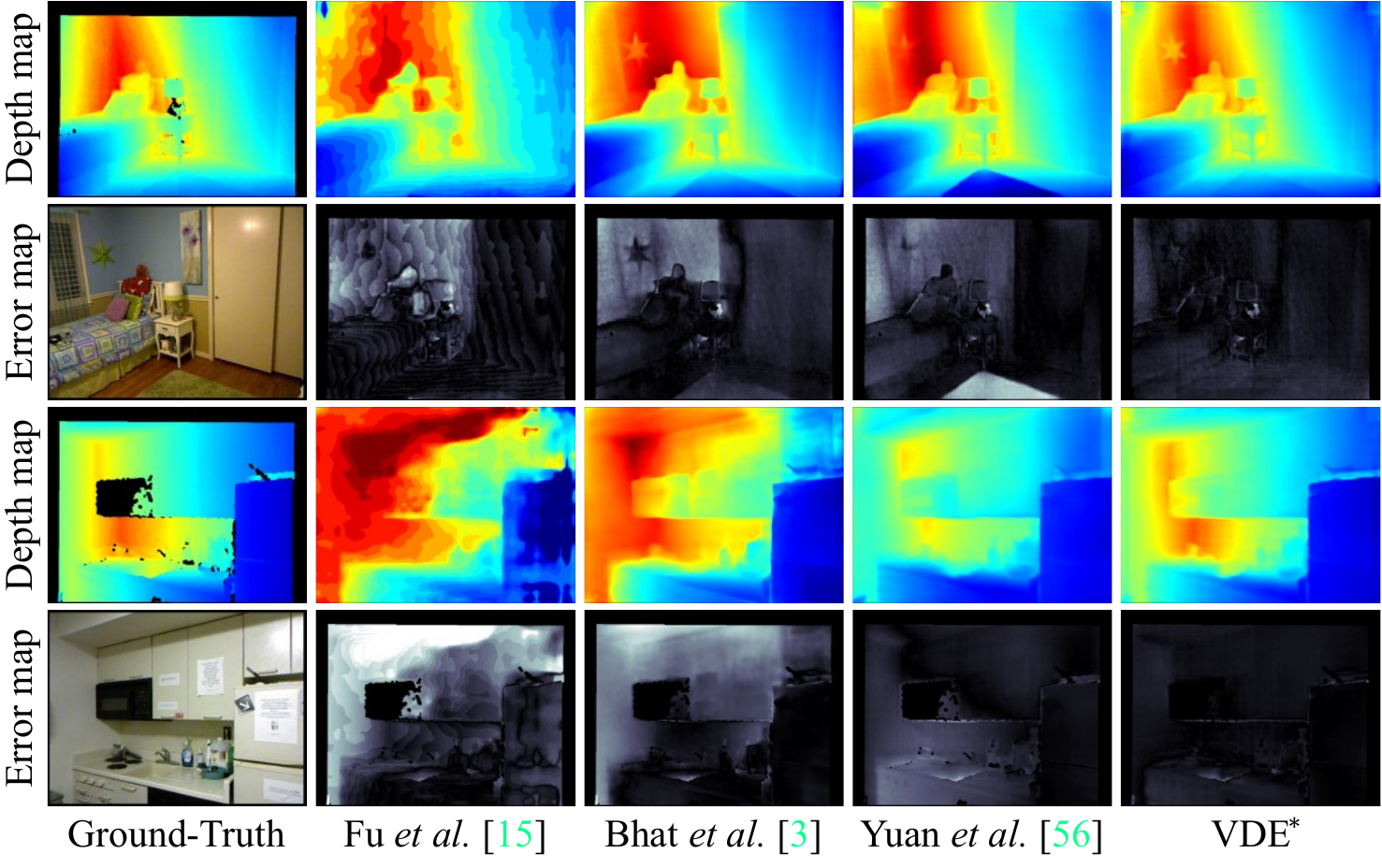}
   \vspace*{-0.4cm}
   \caption{Qualitative results on NYUv2. For each depth map, the error map is provided with brighter pixels indicating larger errors.}
   \label{fig:qualitative_NYU}
   \vspace*{-0.2cm}
\end{figure}

For ordinary depth estimation, one VDE model with a single R2MC is trained only on the NYUv2 dataset for 30 epochs. We use Swin-Large \cite{liu2021swin} as the encoder for a fair comparison with existing state-of-the-art methods \cite{yuan2022newcrfs, agarwal2023attention}. The proposed VDE surpasses all conventional methods in four out of six metrics. Furthermore, VDE$^*$ using extra training data from Table \ref{tb:cameras} for the CRDE provides an additional performance boost. This indicates that training a more reliable CRDE using different cameras can further improve the depth estimation performance for an individual camera.
Note that the same observation can be made from Table \ref{tb:performance_versatile}: VDE performs better than `Separate networks.' Figure \ref{fig:qualitative_NYU} compares qualitative results of VDE$^*$ with those of conventional algorithms \cite{fu2018deep, bhat2021adabins, yuan2022newcrfs}, confirming that VDE$^*$ estimates depths more accurately with fewer artifacts. Metric depth estimation results on the KITTI dataset \cite{geiger2012we} are in the appendix (Section D).

\begin{table}[!t]
    \scriptsize
    \renewcommand{\arraystretch}{1.1}
    \caption{Comparison of depth estimation results on NYUv2. For each metric, the best result is \textbf{boldfaced}, and the second best is \underline{underscored}. $*$ means that additional data is used for training.}
    % \vspace*{-0.2cm}
    \centering
    \begin{tabular}{l|c@{\hskip 0.3em}c@{\hskip 0.3em}c@{\hskip 0.3em}|c@{\hskip 0.7em}c@{\hskip 0.7em}c}
    \toprule
    \multirow{1}{*}{Method} & RMSE$(\downarrow)$ & REL$(\downarrow)$ & $\log$$10$$(\downarrow)$ & $\delta_1 (\uparrow)$ & $\delta_2 (\uparrow)$ & $\delta_3 (\uparrow)$\\
    \midrule
    \multicolumn{1}{l|}{Eigen~\etal~\cite{eigen2014depth}} & 0.641 & 0.158 & - & 0.769 & 0.950 & 0.988\\
    \multicolumn{1}{l|}{Wang~\etal~\cite{wang2015towards}$^*$} & 0.745 & 0.220 & 0.094 & 0.605 & 0.890 & 0.970\\
    \multicolumn{1}{l|}{Laina~\etal~\cite{laina2016deeper}} & 0.573 & 0.127 & 0.055 & 0.811 & 0.953 & 0.988\\
    \multicolumn{1}{l|}{Hao~\etal~\cite{hao2018detail}} & 0.555 & 0.127 & 0.053 & 0.841 & 0.966 & 0.991\\
    \multicolumn{1}{l|}{Fu~\etal~\cite{fu2018deep}} & 0.509 & 0.115 & 0.051 & 0.828 & 0.965 & 0.992\\
    \multicolumn{1}{l|}{Hu~\etal~\cite{hu2019revisiting}} & 0.530 & 0.115 & 0.050 & 0.866 & 0.975 & 0.993\\
    \multicolumn{1}{l|}{Chen~\etal~\cite{chen2019structure}} & 0.514 & 0.111 & 0.048 & 0.878 & 0.977 & 0.994\\
    \multicolumn{1}{l|}{Ramam.~\etal~\cite{ramamonjisoa2019sharpnet}$^*$} & 0.495 & 0.139 & 0.047 & 0.888 & 0.979 & 0.995\\
    \multicolumn{1}{l|}{Yin~\etal~\cite{yin2019enforcing}} & 0.416 & 0.108 & 0.048 & 0.875 & 0.976 & 0.994\\
    \multicolumn{1}{l|}{Hyunh~\etal~\cite{huynh2020guiding}} & 0.412 & 0.108 & - & 0.882 & 0.980 & 0.996\\
    \multicolumn{1}{l|}{Lee and Kim~\cite{lee2020multi}} & 0.430 & 0.119 & 0.050 & 0.870 & 0.974 & 0.993 \\
    \midrule
    \multicolumn{1}{l|}{Lee~\etal~\cite{lee2019big}} & 0.392 & 0.110 & 0.047 & 0.885 & 0.978 & 0.994\\
    \multicolumn{1}{l|}{Bhat~\etal~\cite{bhat2021adabins}} & 0.364 & 0.103 & 0.044& 0.903 & 0.984 & \underline{0.997} \\
    \multicolumn{1}{l|}{Yang~\etal~\cite{yang2021transformer}} & 0.365 & 0.106 & 0.045 & 0.900 & 0.983 & 0.996\\
    \multicolumn{1}{l|}{Ranftl~\etal~\cite{ranftl2021vision}$^*$} & 0.357 & 0.110 & 0.045 & 0.904 & 0.988 & \textbf{0.998} \\
    % \multicolumn{1}{l|}{Jun~\etal~\cite{jun2022depth}$^*$} & 0.362 & 0.100 & 0.043 & 0.907 & 0.986 & \underline{0.997} \\
    \multicolumn{1}{l|}{Jun~\etal~\cite{jun2022depth}$^*$} & 0.355 & 0.098 & 0.042 & 0.913 & 0.987 & \textbf{0.998} \\
    \multicolumn{1}{l|}{Patil~\etal~\cite{patil2022p3depth}} & 0.356 & 0.104 & 0.043 & 0.898 & 0.981 & 0.996 \\
    \multicolumn{1}{l|}{Yuan~\etal~\cite{yuan2022newcrfs}} & 0.334 & 0.095 & 0.041 & 0.922 & \underline{0.992} & \textbf{0.998}\\
    \multicolumn{1}{l|}{Agar. and Arora~\cite{agarwal2023attention}} & 0.322 & 0.090 & 0.039 & 0.929 & 0.991 & \textbf{0.998}\\
    % \multicolumn{1}{l|}{Zhang~\etal~\cite{zhang2023improving}} & 0.321 & 0.089 & 0.039 & 0.932 & - & -\\
    % VDE & \underline{0.325} & \underline{0.091} & \underline{0.039} & \underline{0.926} & 0.990 & \underline{0.997}\\
    % VDE-L & \textbf{0.315} & \textbf{0.088} & \textbf{0.038} & \textbf{0.934} & \underline{0.991} & \textbf{0.998}\\
    % VDE & 0.325 & 0.091 & 0.039 & 0.926 & 0.990 & \underline{0.997}\\
    % VDE-L & \underline{0.315} & \underline{0.088} & \underline{0.038} & \underline{0.934} & \underline{0.991} & \textbf{0.998}\\
    % \midrule
    % VDE & \underline{0.325} & \underline{0.091} & \underline{0.039} & \underline{0.926} & 0.990 & \underline{0.997}\\
    % VDE$^*$ & \textbf{0.307} & \textbf{0.085} & \textbf{0.037} & \textbf{0.941} & \underline{0.991} & \textbf{0.998}\\
    VDE & \underline{0.315} & \underline{0.088} & \underline{0.038} & \underline{0.934} & 0.991 & \textbf{0.998}\\
    VDE$^*$ & \textbf{0.290} & \textbf{0.080} & \textbf{0.035} & \textbf{0.949} & \textbf{0.993} & \textbf{0.998}\\
    \bottomrule
    \end{tabular}
    \label{tb:performance_NYU}
    \vspace*{-0.1cm}
\end{table}

\subsection{Analysis}
To evaluate the generalization performance on unseen data, we compare the proposed CRDE with existing relative depth estimators MiDaS \cite{ranftl2020} and DPT-Large \cite{ranftl2021vision}, which do not use NYUv2 for their training. Table \ref{tb:performance_relative} compares relative $\delta_1$ and Kendall's $\tau$ results on NYUv2. In this test, we train the CRDE on the datasets in Table \ref{tb:cameras}, excluding NYUv2 and SUN (Kinect v1) captured with Kinect v1. The CRDE performs well on the unseen images as well, for it is trained to estimate normalized (\ie relative) depths reliably for diverse datasets. Hence, it surpasses the existing estimators.

\begin{table}[!t]
    \scriptsize
    \setlength{\tabcolsep}{11.5pt}
    \caption{Comparison of relative $\delta_1$ and Kendall's $\tau$ results on NYUv2.}
    % \vspace*{-0.2cm}
    \centering
    \begin{tabular}{l|c|cc}
    \toprule
    & \# Params & Relative $\delta_1(\uparrow)$ & Kendall's $\tau(\uparrow)$\\
    \midrule
    % \# data & & &\\
    MiDaS \cite{ranftl2020} & 105.4M & 0.894 & 0.746\\
    DPT \cite{ranftl2021vision} & 344.1M & 0.905 & 0.764\\
    CRDE & 149.8M & \textbf{0.924} & \textbf{0.843}\\
    \bottomrule
    \end{tabular}
    \label{tb:performance_relative}
\end{table}

\begin{table}[!t]
    \scriptsize
    \setlength{\tabcolsep}{10.0pt}
    \caption{Ablation studies of VDE on NYUv2.}
    \centering
    \begin{tabular}{lcc|cccc}
    \toprule
    \multicolumn{3}{c|}{Settings} & RMSE$(\downarrow)$ & REL$(\downarrow)$ & $\delta_1(\uparrow)$ & $\tau(\uparrow)$\\
    \midrule
    \multicolumn{3}{l|}{Baseline} & 0.335 & 0.094 & 0.922 & 0.844\\
    % \multicolumn{3}{l|}{+ skip-connection} & 0.331 & 0.094 & 0.925 & 0.847\\
    % \multicolumn{3}{l|}{+ normalization} & 0.333 & 0.092 & \textbf{0.926} & 0.846\\
    % \multicolumn{3}{l|}{+ learnable $\alpha, \beta, \gamma$} & 0.330 & 0.092 & \textbf{0.926} & 0.848\\
    \multicolumn{3}{l|}{+ FMM} & 0.330 & 0.092 & \textbf{0.926} & 0.848\\
    \multicolumn{3}{l|}{+ R2MC} & 0.329 & 0.093 & 0.924 & 0.854\\
    \multicolumn{3}{l|}{+ FMM + R2MC} & \textbf{0.325} & \textbf{0.091} & \textbf{0.926} & \textbf{0.858}\\
    \bottomrule
    \end{tabular}
    \label{tb:ablation}
\end{table}

\begin{table}[!t]
    \scriptsize
    \setlength{\tabcolsep}{8.7pt}
    \caption{Impacts of hyperparameters in FMMs. NYUv2 is used in this test.}
    % \vspace*{-0.2cm}
    \centering
    \begin{tabular}{ccc|cccc}
    \toprule
    $\alpha$ & $\beta$ & $\gamma$ & RMSE$(\downarrow)$ & REL$(\downarrow)$ & $\delta_1(\uparrow)$ & $\tau(\uparrow)$\\
    \midrule
    0 & 0 & 0 & 0.335 & 0.094 & 0.922 & 0.844\\
    1 & 1 & 0 & 0.331 & 0.093 & 0.923 & 0.847\\
    0 & 0 & 1 & 0.341 & 0.094 & 0.920 & 0.841\\
    0.5 & 0.5 & 0.5 & 0.332 & 0.093 & 0.924 & \textbf{0.849}\\
    \midrule
    \multicolumn{3}{c|}{Learnable} & \textbf{0.330} & \textbf{0.092} & \textbf{0.926} & 0.848\\
    \bottomrule
    \end{tabular}
    \label{tb:ablation_coeff}
\end{table}

In Table \ref{tb:ablation}, we test ablated methods of VDE on NYUv2. `Baseline' is a basic encoder-decoder network using ordinary Swin transformer blocks instead of FMMs. It estimates metric depths directly, instead of normalized depths, with no R2MC. We see that both FMMs and R2MC lead to performance improvements.

Table \ref{tb:ablation_coeff} compares the results when learnable mixing coefficients $\alpha, \beta, \gamma$ in FMMs are fixed as follows:
\vspace*{-0.2cm}
\begin{enumerate}\itemsep0mm
\item The decoder feature is not mixed with the encoder feature ($\alpha=0, \beta=0, \gamma=0$).
\vspace*{-0.1cm}
\item Queries and keys are obtained from the encoder feature, while values from the decoder feature, as in \cite{yuan2022newcrfs} ($\alpha=1, \beta=1, \gamma=0$).
\vspace*{-0.1cm}
\item The opposite to case 2 ($\alpha=0, \beta=0, \gamma=1$).
\vspace*{-0.1cm}
\item The encoder and decoder features are equally mixed ($\alpha=0.5, \beta=0.5, \gamma=0.5$).
\end{enumerate}
\vspace*{-0.2cm}
Without testing all combinations, `Learnable' yields the best results. More detailed analyses of FMMs and R2MCs are conducted in the appendix (Section F).

\section{Conclusions}
\label{sec:conclusions}
We proposed a novel VDE composed of a single CRDE and multiple R2MCs. First, the CRDE generates common relative depth features. Then, the R2MCs convert the relative features to camera-specific metric depth maps. It was shown that VDE is applicable to versatile depth estimation successfully, as well as providing state-of-the-art performance in the single-camera scenario. More specifically, VDE yields comparable or better performance than non-versatile techniques optimized for specific cameras, while requiring only a 1.12\% parameter increase per camera.

%%%%%%%%% REFERENCES
{\small
\bibliographystyle{ieee_fullname}
\bibliography{2023_ICCV_JYJUN}
}
\clearpage
\onecolumn
\begin{appendices}
\renewcommand{\thesection}{\Alph{section}}

\section{Network Architecture}\label{network_archietcture}
\begin{figure*}[!h]
  \centering
  \includegraphics[width=\linewidth]{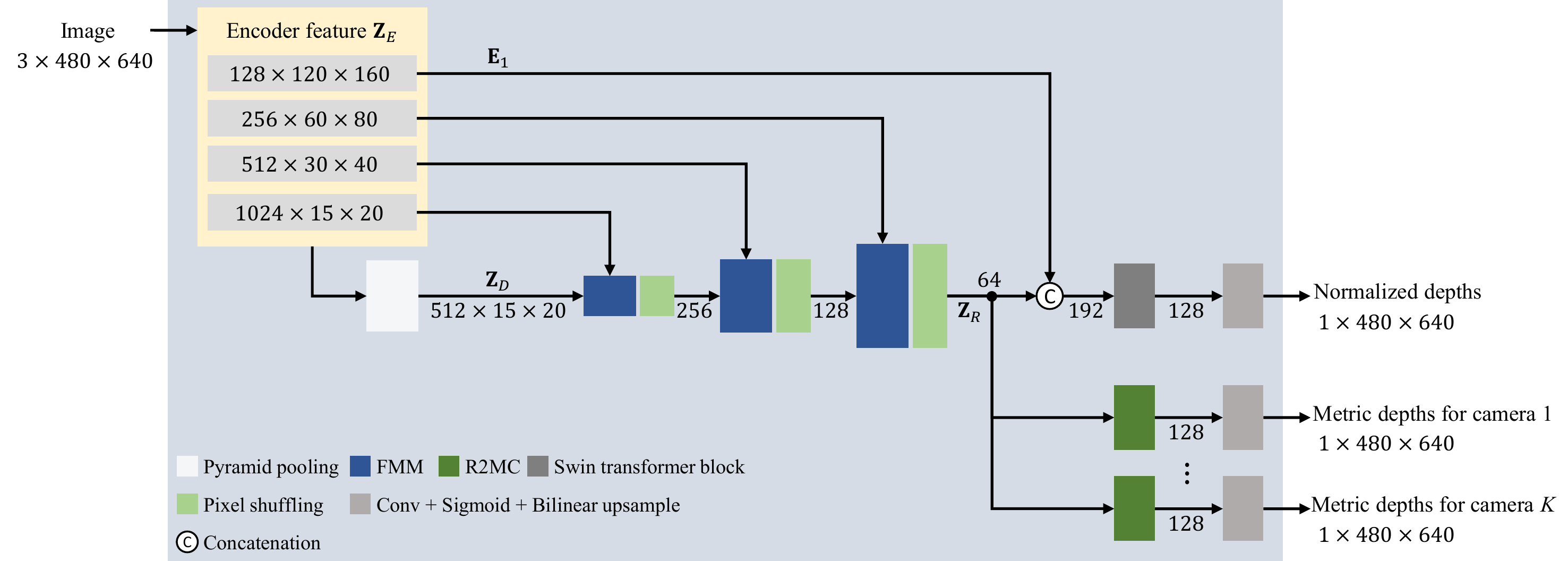}
  \caption{Network structure of the proposed VDE. The number of channels in each feature tensor is specified.}
  \label{fig:detailed_architecture}
\end{figure*}

Figure~\ref{fig:detailed_architecture} shows the network structure of the proposed VDE. As the encoder, we adopt Swin-Base \cite{liu2021swin}, which processes a $480 \times 640$ RGB image to yield the multi-scale encoder feature $\bZ_E$. Then, we compress $\bZ_E$  via pyramid pooling \cite{zhao2017pyramid} to generate the decoder feature $\bZ_D$ of size $512 \times 15 \times 20$. Next, we process $\bZ_D$ sequentially using three FMMs with pixel shuffling layers to yield the relative depth feature $\bZ_R$ of size $64 \times 120 \times 160$.
We then concatenate $\bZ_R$ with $\bE_1$ from the encoder to perform the normalized depth estimation. Also, we feed $\bZ_R$ to $K$ R2MCs to generate metric depths for the corresponding cameras. The detailed structures of FMMs and R2MCs are in Figure~\ref{fig:FMM_ML} and Figure~\ref{fig:R2MC_CL}, respectively.

\vspace{1cm}
\clearpage

\section{Dataset Specifications}\label{dataset_specification}

Table \ref{tb:dataset_specification} summarizes the camera specifications and the training and test images for each dataset. For NYUv2, we adopt the training split in \cite{bhat2021adabins, lee2019big, yuan2022newcrfs}. For ScanNet, we sample every 100th and 200th frames from the official training and test scans, respectively. For versatile depth estimation, we sample the training images in a dataset uniformly to construct each sub-dataset of 1,000 images, except for SUN (RealSense) for which we use all 587 images available. SUN RGB-D includes Kinect v1 and Kinect v2 images. However, these images have different characteristics from NYUv2 and DIML-Indoor ones because SUN RGB-D performs post-processing differently to improve the depth maps. We also conduct additional experiments using KITTI and HR-WSI. We adopt the Eigen split \cite{eigen2015predicting} of KITTI and the official split of HR-WSI.

\vspace*{1cm}
\begin{table*}[!h]
    \small
    \setlength{\tabcolsep}{8.0pt}
    \caption{Dataset specifications}
    \centering
    \begin{tabular}{l|c|ccc|cc}
    \toprule
    Dataset & Scene type & Camera & Field of view & Depth range & Training images & Test images\\
    \midrule
    NYUv2 \cite{silberman2012indoor} & Indoor & Kinect v1 & 43\textdegree $\times$ 57\textdegree & 0.4$\sim$4.0m & 36,253 & 654\\
    \midrule
    \multirow{2}{*}{DIML \cite{kim2018deep}} & Indoor & Kinect v2 & 60\textdegree $\times$ 70\textdegree & 0.5$\sim$8.0m & 1,609 & 504\\
    & Outdoor & ZED stereo & 60\textdegree $\times$ 90\textdegree & 0.5$\sim$20m & 1,505 & 500\\
    \midrule
    \multirow{2}{*}{DIODE \cite{vasiljevic2019diode}} & Indoor & \multirow{2}{*}{Laser scanner} & \multirow{2}{*}{305\textdegree $\times$ 360\textdegree} & \multirow{2}{*}{0.6$\sim$350m} & 8,574 & 325\\
    & Outdoor & & & & 16,884 & 446\\
    \midrule
    ScanNet \cite{dai2017scannet} & Indoor & Structure sensor & 45\textdegree $\times$ 58\textdegree & 0.4$\sim$3.5m & 25,536 & 1,084\\
    \midrule
    \multirow{4}{*}{SUN RGB-D \cite{song2015sun}} & \multirow{4}{*}{Indoor} & Kinect v1 & 43\textdegree $\times$ 57\textdegree & 0.4$\sim$4.0m & 1,073 & 930\\
    & & Kinect v2 & 60\textdegree $\times$ 70\textdegree & 0.5$\sim$8.0m & 1,924 & 1,860\\
    & & RealSense & 43\textdegree $\times$ 70\textdegree & 0.4$\sim$2.8m & 587 & 572\\
    & & Xtion & 45\textdegree $\times$ 58\textdegree & 0.8$\sim$3.5m & 1,701 & 1,688\\
    \midrule
    KITTI \cite{geiger2012we} & Outdoor & HDL-64E & 26.8\textdegree $\times$ 360\textdegree & 2.5$\sim$80m & 23,158 & 697\\
    \midrule
    HR-WSI \cite{xian2020structure} & Indoor \& Outdoor & \multicolumn{3}{|c|}{Stereo image pairs} & 20,378 & 400\\
    \bottomrule
    \end{tabular}
    \label{tb:dataset_specification}
\end{table*}
\clearpage

\section{Versatile Depth Estimation}\label{versatile_depth_estimation_results}
\subsection{Extended VDE test}
\label{ssec:versatile_performance_single_network}

Table \ref{tb:performance_versatile_ext} is an extended version of Table \ref{tb:performance_versatile}, including the results of `Single network.' Also, for an easier comparison, we report the geometric mean scores over the 10 sub-datasets.

\begin{table}[!h]
    \small
    \setlength{\tabcolsep}{6.7pt}
    \caption{Comparison of versatile depth estimation results. For each setting, the required number of parameters is specified. Lower RMSE and REL indicate better results, while higher $\delta_1$ and $\tau$ are better ones. In each test, the best result is \textbf{boldfaced}.}
    \centering
    \begin{tabular}{c|l|ccccc}
    \toprule
    Setting & \multicolumn{1}{|c|}{Dataset} & RMSE$(\downarrow)$ & REL$(\downarrow)$ & $\delta_1(\uparrow)$ & $\tau(\uparrow)$\\
    \midrule
    \multirow{11}{*}{\shortstack{Single\\network\\(149.8M)}} & NYUv2 & 0.413 & 0.121 & 0.862 & 0.813\\
    & DIML-Indoor & 0.916 & 0.333 & 0.478 & 0.573\\
    & DIML-Outdoor & 3.172 & 0.373 & 0.328 & 0.693\\
    & DIODE-Indoor & 2.607 & 0.462 & 0.123 & 0.479\\
    & DIODE-Outdoor & 15.647 & 0.797 & 0.005 & 0.574\\
    & ScanNet & 0.414 & 0.225 & 0.658 & 0.719\\
    & SUN (Kinect v1) & 0.402 & 0.227 & 0.672 & 0.764\\
    & SUN (Kinect v2) & 0.504 & 0.275 & 0.517 & 0.811\\
    & SUN (RealsSense) & 0.307 & 0.162 & 0.795 & 0.792\\
    & SUN (Xtion) & 0.513 & 0.551 & 0.611 & 0.731\\
    \cmidrule{2-6}
    & \multicolumn{1}{|c|}{Mean} & 0.957 & 0.305 & 0.314 & 0.685\\
    \midrule
    \multirow{11}{*}{\shortstack{Separate\\networks\\(149.8M \!\! $\times$ \!\! 10)}} & NYUv2 & 0.376 & 0.111 & 0.892 & 0.811\\
    & DIML-Indoor & 0.693 & 0.244 & 0.640 & 0.585\\
    & DIML-Outdoor & \textbf{1.240} & \textbf{0.142} & 0.812 & 0.828\\
    & DIODE-Indoor & 1.352 & 0.255 & 0.569 & 0.543\\
    & DIODE-Outdoor & \textbf{5.880} & \textbf{0.302} & \textbf{0.615} & \textbf{0.646}\\
    & ScanNet & 0.347 & 0.169 & 0.774 & 0.691\\
    & SUN (Kinect v1) & 0.324 & 0.157 & 0.803 & 0.754\\
    & SUN (Kinect v2) & 0.257 & 0.099 & 0.912 & 0.829\\
    & SUN (RealsSense) & 0.245 & 0.104 & 0.893 & 0.782\\
    & SUN (Xtion) & 0.405 & 0.416 & 0.784 & 0.720\\
    \cmidrule{2-6}
    & \multicolumn{1}{|c|}{Mean} & 0.612 & 0.179 & 0.760 & 0.712\\
    \midrule
    \multirow{11}{*}{\shortstack{Multiple\\decoders\\(571.3M)}} & NYUv2 & 0.380 & 0.109 & 0.894 & 0.817\\
    & DIML-Indoor & 0.662 & 0.235 & 0.673 & 0.616\\
    & DIML-Outdoor & 1.356 & 0.148 & 0.800 & 0.838\\
    & DIODE-Indoor & 1.180 & \textbf{0.217} & 0.638 & 0.644\\
    & DIODE-Outdoor & 6.973 & 0.351 & 0.485 & 0.633\\
    & ScanNet & 0.340 & 0.185 & 0.753 & 0.739\\
    & SUN (Kinect v1) & 0.349 & 0.184 & 0.753 & 0.770\\
    & SUN (Kinect v2) & 0.330 & 0.168 & 0.797 & 0.827\\
    & SUN (RealsSense) & 0.223 & 0.110 & 0.906 & 0.799\\
    & SUN (Xtion) & 0.417 & 0.467 & 0.748 & 0.745\\
    \cmidrule{2-6}
    & \multicolumn{1}{|c|}{Mean} & 0.632 & 0.196 & 0.734 & 0.738\\
    \midrule
    \multirow{11}{*}{\shortstack{VDE\\(167.3M)}} & NYUv2 & \textbf{0.335} & \textbf{0.093} & \textbf{0.925} & \textbf{0.848}\\
    & DIML-Indoor & \textbf{0.653} & \textbf{0.228} & \textbf{0.678} & \textbf{0.635}\\
    & DIML-Outdoor & 1.245 & 0.146 & \textbf{0.846} & \textbf{0.853}\\
    & DIODE-Indoor & \textbf{1.175} & 0.222 & \textbf{0.645} & \textbf{0.693}\\
    & DIODE-Outdoor & 6.141 & 0.334 & 0.566 & 0.642\\
    & ScanNet & \textbf{0.295} & \textbf{0.139} & \textbf{0.830} & \textbf{0.778}\\
    & SUN (Kinect v1) & \textbf{0.287} & \textbf{0.125} & \textbf{0.863} & \textbf{0.804}\\
    & SUN (Kinect v2) & \textbf{0.231} & \textbf{0.088} & \textbf{0.939} & \textbf{0.857}\\
    & SUN (RealsSense) & \textbf{0.215} & \textbf{0.098} & \textbf{0.925} & \textbf{0.832}\\
    & SUN (Xtion) & \textbf{0.363} & \textbf{0.411} & \textbf{0.838} & \textbf{0.784}\\
    \cmidrule{2-6}
    & \multicolumn{1}{|c|}{Mean} & \textbf{0.559} & \textbf{0.164} & \textbf{0.795} & \textbf{0.768}\\
    \bottomrule
    \end{tabular}
    \label{tb:performance_versatile_ext}
\end{table}
\clearpage

\subsection{Cross-dataset evaluation}\label{ssec:cross_dataset_eval}

Note that each R2MC is optimized for a specific camera. In this test, we conduct cross-dataset evaluation, which assesses the relative depth estimation results of each R2MC on different datasets obtained with different cameras. We measure Kendall's $\tau$ scores, since the metrics for metric depth estimation are inappropriate for this evaluation across different cameras.

\begin{figure*}[h]
  \centering
  \includegraphics[width=0.93\linewidth]{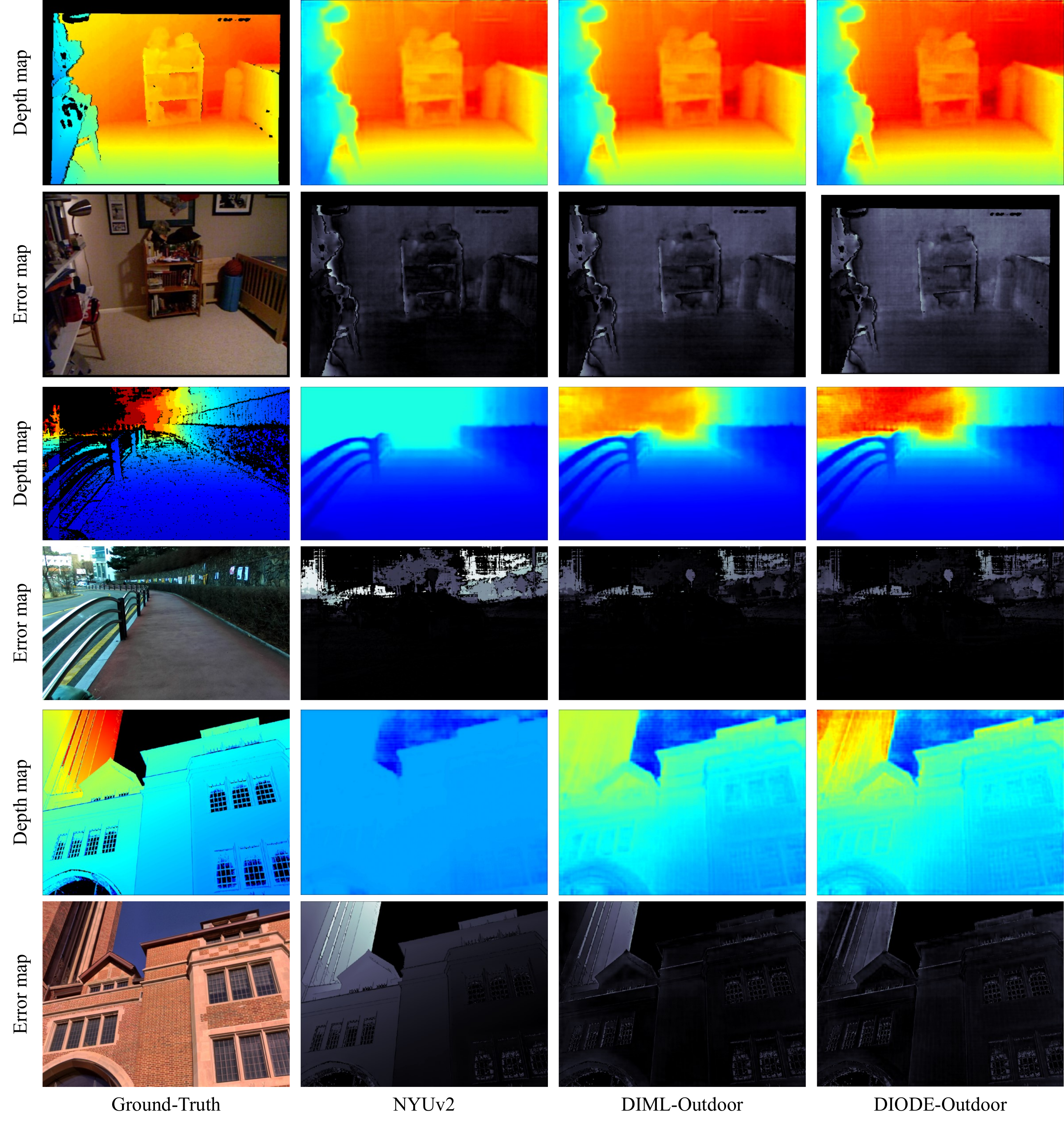}
   \caption{For each input image, three depth maps, obtained by the R2MCS for NYUv2, DIML-Outdoor, and DIODE-Outdoor, are provided with the corresponding error maps. From top to bottom, the input images are from the NYUv2, DIML-Outdoor, and DIODE-Outdoor test datasets.}
   \label{fig:camera_range_difference}
\end{figure*}

First, Figure \ref{fig:camera_range_difference} compares depth maps obtained by the proposed R2MCs for NYUv2, DIML-Outdoor, and DIODE-Outdoor for the same input images. It can be observed that, for each input image, the three depth maps contain similar relative depth information even though their depth scales are different. In other words, each R2MC predicts relative depth orders reliably even for images captured with different cameras.

\begin{figure*}[h]
  \centering
  \includegraphics[width=\linewidth]{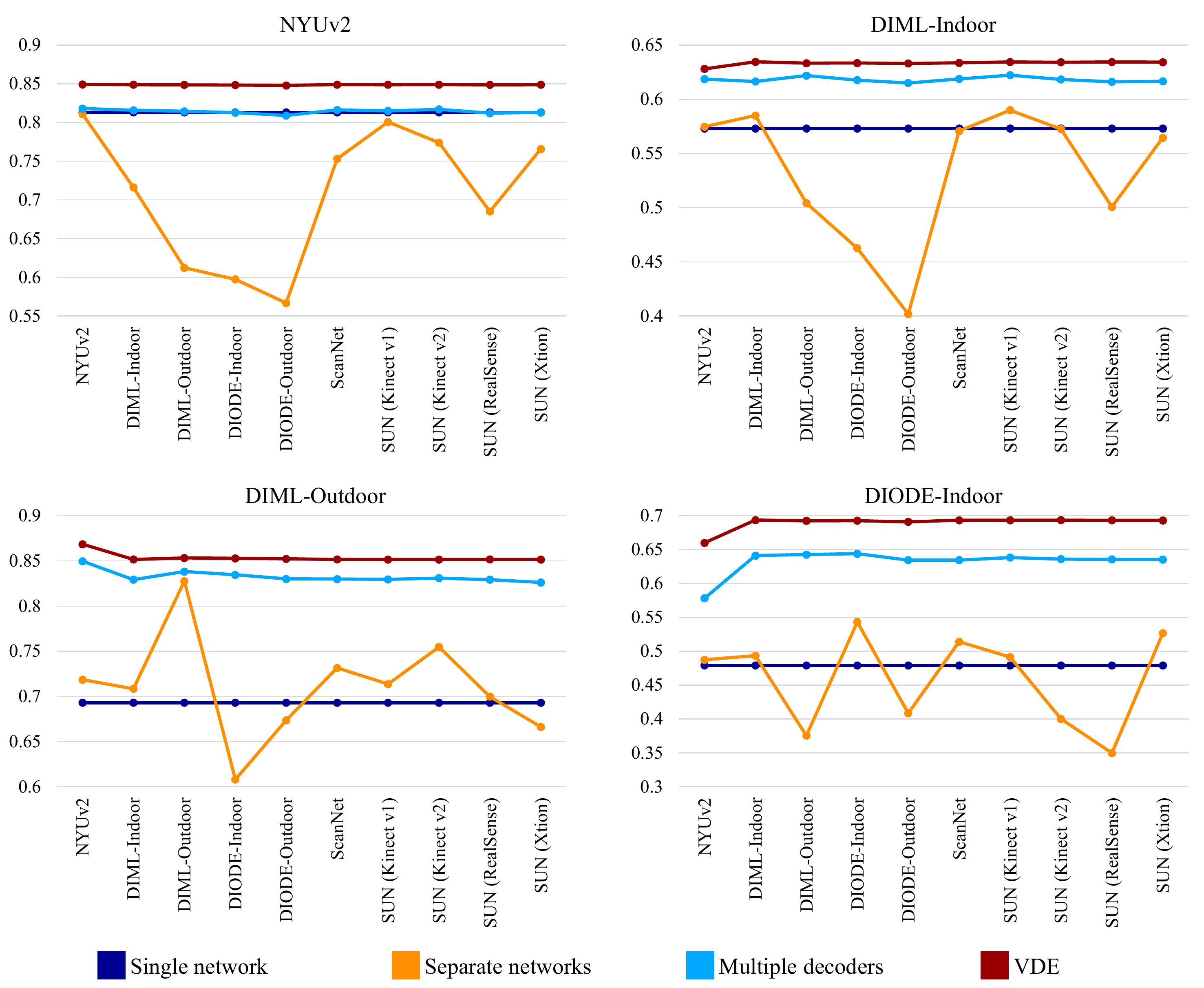}
   \caption{Comparison of Kendall's $\tau$ scores in the cross-dataset evaluation.}
   \label{fig:cross_eval_performance}
\end{figure*}

\begin{figure*}[h]
  \centering
  \includegraphics[width=\linewidth]{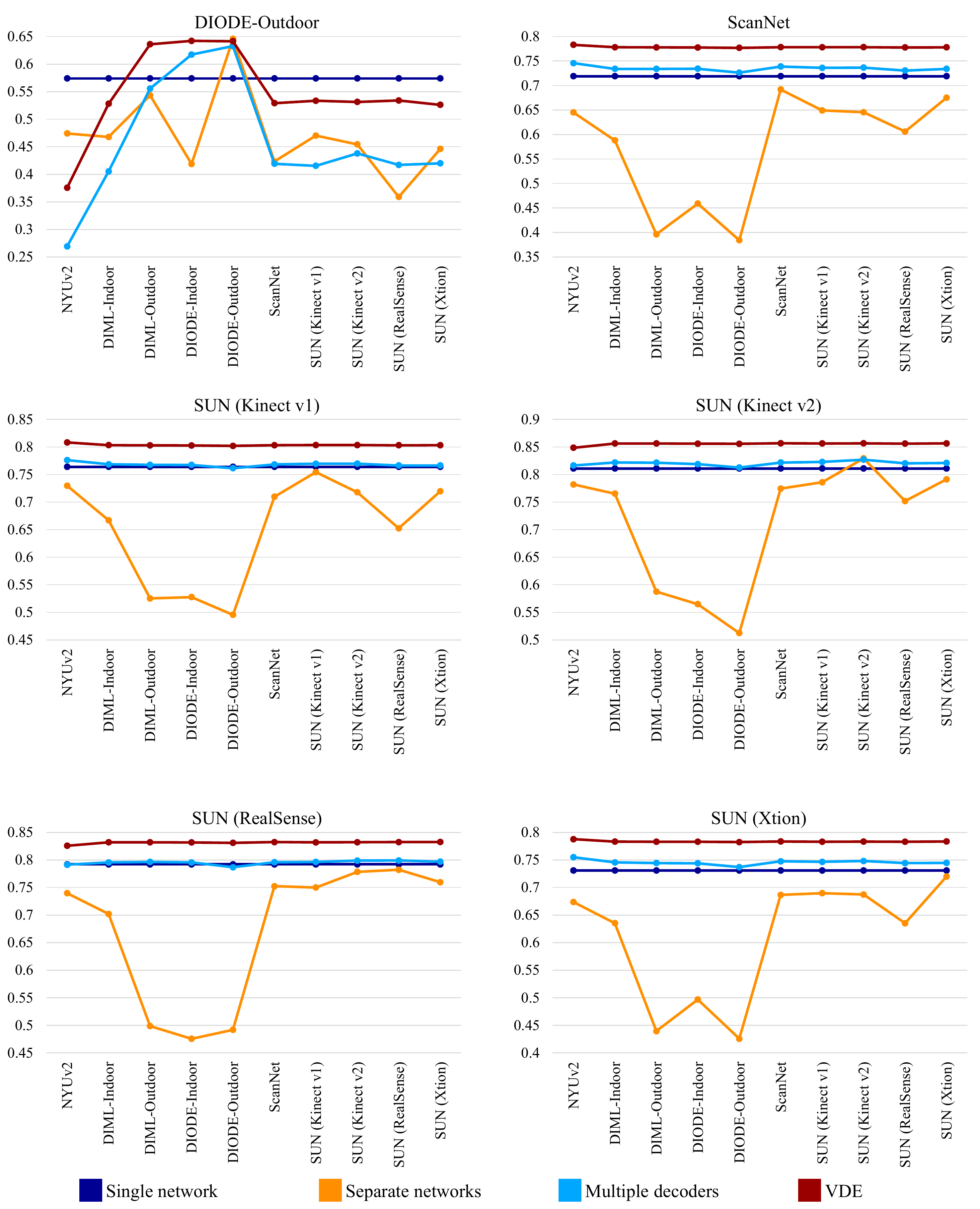}
   \caption{Comparison of Kendall's $\tau$ scores in the cross-dataset evaluation.}
   \label{fig:cross_eval_performance2}
\end{figure*}

Next, Figure \ref{fig:cross_eval_performance} and Figure \ref{fig:cross_eval_performance2} compare Kendall's $\tau$ scores in this cross-dataset evaluation. For example, the top left graph in Figure \ref{fig:cross_eval_performance} compares the Kendall's $\tau$ scores on the NYUv2 test images obtained by the 10 R2MCs for the 10 sub-datasets. Since only a single score is generated by `Single network,' it is repeated for the 10 R2MCs. We see that `Separate networks' are fitted to the corresponding sub-datasets and perform poorly on the other datasets in general, while `Multiple decoders' and VDE provide consistent Kendall's $\tau$ scores. However, VDE outperforms `Multiple decoders,' indicating that sharing more portions of the network for the common goal of normalized depth estimation via CRDE improves the generalization performance. In the main paper, it is shown that VDE surpasses the other settings in metric depth evaluation as well, indicating that R2MC successfully transforms relative depth features into metric depths.

\vspace*{1cm}

\clearpage

\subsection{VDE using multiple datasets of different sizes}\label{ssec:VDE_using_heterogenous_sized_datasets}
In real-world applications of VDE, different datasets may have different sizes. We hence evaluate the performances of VDE using multiple datasets of various sizes in Table \ref{tb:performance_versatile_NYU_DIML}. In this experiment, we use the whole training datasets of NYUv2 and DIML, which are specified in Table \ref{tb:dataset_specification}. Since NYUv2 contains much more training images than DIML, each DIML image is used five times per epoch while each NYUv2 image is used only once per epoch during the training. Similarly to Table \ref{tb:performance_versatile}, the proposed VDE outperforms `Separate networks' on indoor images. Especially, note that VDE surpasses `Separate networks' even on NYUv2, for which `Separate network' is optimized with a relatively large number of training images.

\begin{table}[!h]
    \small
    \setlength{\tabcolsep}{4.6pt}
    \caption{Comparison of versatile depth estimation results using uneven sized datasets.}
    \vspace{0.1cm}
    \centering
    \begin{tabular}{c|l|c|ccccc}
    \toprule
    Setting & \multicolumn{1}{c|}{Dataset} & \# data & RMSE$(\downarrow)$ & REL$(\downarrow)$ & $\delta_1(\uparrow)$ & $\tau(\uparrow)$\\
    \midrule
    \multirow{4}{*}{\shortstack{Separate\\networks\\(149.8M \!\! $\times$ \!\! 3)}} & NYUv2 & 36,253 & 0.330 & 0.092 & 0.926 & 0.848\\
    & DIML-Indoor & 1,609 & 0.683 & 0.246 & 0.650 & 0.589\\
    & DIML-Outdoor & 1,505 & \textbf{1.180} & \textbf{0.135} & \textbf{0.830} & 0.840\\
    \cmidrule{2-7}
    & \multicolumn{2}{|c|}{Mean} & 0.643 & \textbf{0.145} & 0.793 & 0.749\\
    % \midrule
    % \multirow{3}{*}{\shortstack{Multiple\\decoders\\(196.7M)}} & NYUv2 &\\
    % & DIML-Indoor & \\
    % & DIML-Outdoor & \\
    \midrule
    \multirow{4}{*}{\shortstack{VDE\\(155.1M)}} & NYUv2 & 36,253 & \textbf{0.322} & \textbf{0.091} & \textbf{0.928} & \textbf{0.859}\\
    & DIML-Indoor & 1,609 & \textbf{0.667} & \textbf{0.239} & \textbf{0.662} & \textbf{0.629}\\
    & DIML-Outdoor & 1,505 & 1.217 & 0.141 & 0.818 & \textbf{0.858}\\
    \cmidrule{2-7}
    & \multicolumn{2}{|c|}{Mean} & \textbf{0.639} & \textbf{0.145} & \textbf{0.795} & \textbf{0.774}\\
    \bottomrule
    \end{tabular}
    \label{tb:performance_versatile_NYU_DIML}
    %\vspace*{-0.1cm}
\end{table}

\vspace*{1cm}
Table \ref{tb:performance_versatile_SUNRGBD} compares the results using the whole sub-datasets of SUN RGB-D. Again, the proposed VDE provides better results than `Separate networks' in most cases.

\begin{table*}[!h]
    \small
    \setlength{\tabcolsep}{6.7pt}
    \caption{Comparison of versatile depth estimation results on the SUN RGB-D dataset.}
    \centering
    \begin{tabular}{c|l|c|cccc}
    \toprule
    Setting & Dataset & \# data & RMSE$(\downarrow)$ & REL $(\downarrow)$ & $\delta_1 (\uparrow)$ & $\tau (\uparrow)$\\
    \midrule
    \multirow{5}{*}{\shortstack{Separate\\networks\\(149.8M \!\! $\times$ \!\! 4)}} & Kinect v1 & 1,073 & 0.325 & 0.153 & 0.805 & 0.736\\
    & Kinect v2 & 1,924 & 0.240 & 0.091 & 0.927 & 0.836\\
    % & RealSense & 0.245 & 0.110 & 0.886 & 0.742\\
    & RealSense & 587 & 0.245 & 0.104 & 0.893 & 0.782\\
    & Xtion & 1,701 & 0.399 & \textbf{0.381} & 0.798 & 0.710\\
    \cmidrule{2-7}
    & \multicolumn{2}{|c|}{Mean} & 0.296 & 0.153 & 0.854 & 0.765\\
    \midrule
    \multirow{5}{*}{\shortstack{VDE\\(156.8M)}} & Kinect v1 & 1,073 & \textbf{0.289} & \textbf{0.132} & \textbf{0.855} & \textbf{0.797}\\
    & Kinect v2 & 1,924 & \textbf{0.220} & \textbf{0.083} & \textbf{0.945} & \textbf{0.860}\\
    & RealSense & 587 & \textbf{0.194} & \textbf{0.085} & \textbf{0.950} & \textbf{0.835}\\
    & Xtion & 1,701 & \textbf{0.339} & 0.397 & \textbf{0.853} & \textbf{0.783}\\
    \cmidrule{2-7}
    & \multicolumn{2}{|c|}{Mean} & \textbf{0.254} & \textbf{0.139} & \textbf{0.900} & \textbf{0.818} \\
    \bottomrule
    \end{tabular}
    \label{tb:performance_versatile_SUNRGBD}
\end{table*}
\clearpage

\section{Ordinary Depth Estimation on KITTI}
\begin{figure*}[!h]
\vspace{-0.4cm}
  \centering
  \includegraphics[width=0.85\linewidth]{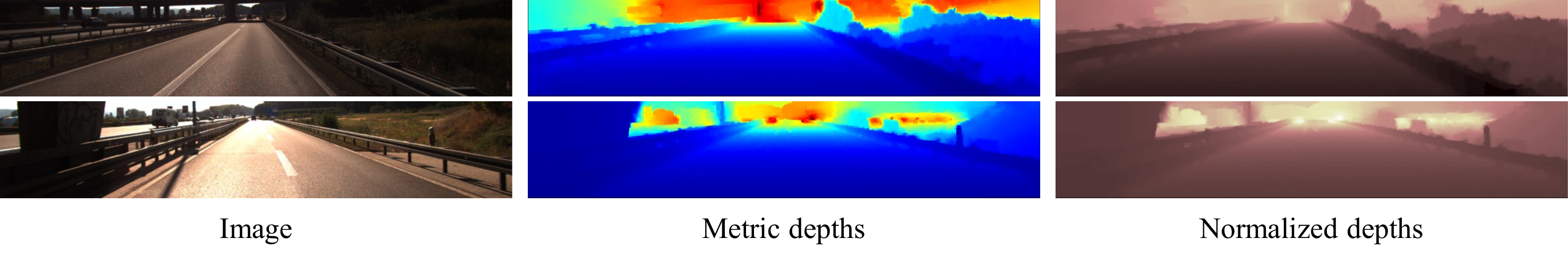}
   \caption{Metric depths and its normalized depths on KITTI.}
   \label{fig:kitti_norm_err}
\vspace{-0.4cm}
\end{figure*}

Since KITTI is collected by a fixed camera on a vehicle, the metric depth values of a certain pixel in different road scenes tend to be almost identical if the corresponding road points are not occluded by other objects, as illustrated in Figure~\ref{fig:kitti_norm_err}. However, the relative depth order of pixels in a road scene can vary significantly due to objects in the scene. In other words, pixels at the same distance may have quite different values after the normalization, as  in Figure~\ref{fig:kitti_norm_err}. Therefore, the proposed VDE, which exploits normalized depth features, does not have an advantage on the KITTI dataset.

\begin{figure*}[!h]
  \centering
  \includegraphics[width=0.95\linewidth]{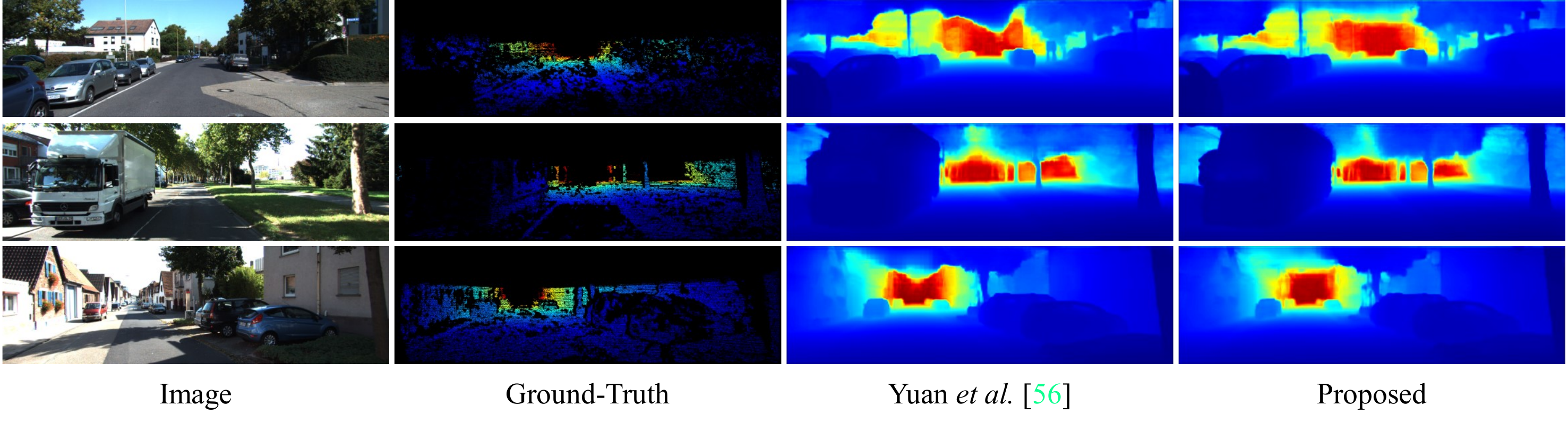}
   \caption{Qualitative comparison on KITTI.}
   \label{fig:qualitative_kitti_supp}
\end{figure*}

Table \ref{tb:tau_performance} compares the Kendall's $\tau$ performances of the proposed VDE (using normalized depths for training) and a modified method (using metric depths for training). We see that on KITTI the modified method provides a better Kendall's $\tau$. This confirms that depth normalization is not helpful on KITTI. Therefore, for KITTI we train the proposed network to yield metric depth maps directly without employing an R2MC. Table \ref{tb:performance_KITTI} compares the performance of this network on the Eigen split of KITTI. We see that the proposed network provides state-of-the-art results on KITTI, which indicates that the proposed network using FMMs is also effective even when no R2MC is employed. Figure \ref{fig:qualitative_kitti_supp} compares estimated depth maps of the proposed network with those of Yuan \etal \cite{yuan2022newcrfs}.

\begin{table*}[!h]
    \small
    \addtolength{\tabcolsep}{15.0pt}
    \caption{Comparison of Kendall's $\tau$ performances on NYUv2 and KITTI.}
    % \vspace*{-0.2cm}
    \centering
    \begin{tabular}{c|ccccc}
    \toprule
    Training & NYUv2 & KITTI\\
    \midrule
    Metric depths & 0.848 & \textbf{0.935}\\
    Normalized depths & \textbf{0.858} & 0.915\\
    \bottomrule
    \end{tabular}
    \label{tb:tau_performance}
\end{table*}

\begin{table}[!h]
    \small
    \addtolength{\tabcolsep}{4.0pt}
    \caption{Comparison of depth estimation results on the Eigen split of KITTI. For each metric, the best result is \textbf{boldfaced}.}
    \centering
    \begin{tabular}{l|ccccccc}
    \toprule
    Method & RMSE$(\downarrow)$ & $\textrm{RMSE}_{\log}(\downarrow)$ & REL$(\downarrow)$ & Sq REL$(\downarrow)$ & $\delta_1(\uparrow)$ & $\delta_2(\uparrow)$ & $\delta_3(\uparrow)$\\
    \midrule
    \midrule
    % \multicolumn{1}{l|}{Eigen~\etal [\textcolor{green}{11}]} & 6.307 & 0.282 & 0.203 & 1.548 & 0.702 & 0.898 & 0.967\\
    % \multicolumn{1}{l|}{Laina~\etal [\textcolor{green}{25}]} & - & - & - & - & - & - & -\\
    % \multicolumn{1}{l|}{Hao~\etal [\textcolor{green}{17}]} & - & - & - & - & - & - & -\\
    \multicolumn{1}{l|}{Fu~\etal \cite{fu2018deep}} & 2.727 & 0.120 & 0.072 & 0.307 & 0.932 & 0.984 & 0.994\\
    % \multicolumn{1}{l|}{Hu~\etal [\textcolor{green}{19}]} & - & - & - & - & - & - & -\\
    \multicolumn{1}{l|}{Lee~\etal \cite{lee2019big}} & 2.756 & 0.096 & 0.059 & 0.245 & 0.956 & 0.993 & 0.998\\
    \multicolumn{1}{l|}{Yin~\etal \cite{yin2019enforcing}} & 3.258 & 0.117 & 0.072 & - & 0.938 & 0.990 & 0.998\\
    \multicolumn{1}{l|}{Lee and Kim \cite{lee2020multi}} & 4.512 & 0.176 & 0.115 & - & 0.864 & 0.962 & 0.986\\
    \multicolumn{1}{l|}{Ranftl~\etal \cite{ranftl2021vision}$^*$} & 2.573 & 0.092 & 0.062 & - & 0.959 & 0.995 & \textbf{0.999}\\
    \multicolumn{1}{l|}{Bhat~\etal \cite{bhat2021adabins}} & 2.360 & 0.088 & 0.058 & 0.190 & 0.964 & 0.995 & \textbf{0.999}\\
    \multicolumn{1}{l|}{Patil~\etal \cite{patil2022p3depth}} & 2.842 & 0.103 & 0.079 & 0.270 & 0.953 & 0.993 & 0.998\\
    \multicolumn{1}{l|}{Yuan~\etal \cite{yuan2022newcrfs}} & 2.129 & 0.079 & 0.052 & 0.155 & 0.974 & \textbf{0.997} & \textbf{0.999}\\
    \midrule
    \multicolumn{1}{l|}{Proposed} & \textbf{2.070} & \textbf{0.078} & \textbf{0.051} & \textbf{0.148} & \textbf{0.975} & \textbf{0.997} & \textbf{0.999}\\

    \bottomrule
    \end{tabular}
    \label{tb:performance_KITTI}
\end{table}

\vspace{-0.3cm}
\clearpage

\section{Relative Depth Estimation}
In Table \ref{tb:performance_relative}, we use the trained parameters of \cite{ranftl2021vision} and \cite{ranftl2020} from their official repositories to evaluate the generalization performance. These models use much more training images than the proposed CRDE --- about 1.4M training images. For a fair comparison, we retrain \cite{ranftl2020, ranftl2021vision}, and the proposed CRDE using the HR-WSI dataset \cite{xian2020structure}, which is a relative depth dataset containing 20,378 training images. Also, we define another metric, `Relative RMSE,' which computes the RMSE score between the ground truth depth map $\bD$ and a calibrated depth map $\Tilde{\bR}$, similarly to `Relative $\delta_1$.' Tables~\ref{tb:performance_relative_NYU} and \ref{tb:performance_relative_KITTI} compare relative RMSE, relative $\delta_1$, and Kendall's $\tau$ results on NYUv2 and KITTI, respectively. We see that the proposed CRDE also outperforms \cite{ranftl2021vision} and \cite{ranftl2020} when trained with the same data.

\vspace*{1cm}
\begin{table}[!h]
    \small
    \setlength{\tabcolsep}{11.5pt}
    \caption{Comparison of relative RMSE, relative $\delta_1$ and Kendall's $\tau$ results on NYUv2.}
    % \vspace*{-0.2cm}
    \centering
    \begin{tabular}{l|c|ccc}
    \toprule
    & \# Params & Relative RMSE$(\downarrow)$ & Relative $\delta_1(\uparrow)$ & Kendall's $\tau(\uparrow)$\\
    \midrule
    % \# data & & &\\
    MiDaS \cite{ranftl2020} & 105.4M & 0.542 & 0.805 & 0.657\\
    DPT \cite{ranftl2021vision} & 344.1M & 0.539 & 0.828 & 0.712\\
    % CRDE (w/o normaliation) & 149.8M & 0.843 & 0.706\\
    CRDE & 149.8M & \textbf{0.455} & \textbf{0.874} & \textbf{0.737}\\
    \bottomrule
    \end{tabular}
    \label{tb:performance_relative_NYU}
\end{table}

\vspace*{1cm}
\begin{table}[!h]
    \small
    \setlength{\tabcolsep}{11.5pt}
    \caption{Comparison of relative RMSE, relative $\delta_1$ and Kendall's $\tau$ results on KITTI.}
    % \vspace*{-0.2cm}
    \centering
    \begin{tabular}{l|c|ccc}
    \toprule
    & \# Params & Relative RMSE$(\downarrow)$ & Relative $\delta_1(\uparrow)$ & Kendall's $\tau(\uparrow)$\\
    \midrule
    % \# data & & &\\
    MiDaS \cite{ranftl2020} & 105.4M & 7.779 & 0.650 & 0.695\\
    DPT \cite{ranftl2021vision} & 344.1M & 5.248 & 0.831 & 0.818\\
    % CRDE (w/o normaliation) & 149.8M & 0.858 & 0.833\\
    CRDE & 149.8M & \textbf{5.060} & \textbf{0.862} & \textbf{0.840}\\
    \bottomrule
    \end{tabular}
    \label{tb:performance_relative_KITTI}
\end{table}
\clearpage

\section{More Analysis on FMM and R2MC}
\subsection{FMM}
Table \ref{tb:ablation_FMM_supp} reports the results of an ablation study on FMM components. `Baseline' is the same as the one in Table \ref{tb:ablation}, which is a basic encoder-decoder network using Swin transformer blocks. Note that each FMM component improves depth estimation results. Table \ref{tb:ablation_coeff_supp} is an extended version of Table  \ref{tb:ablation_coeff}, which compares different combinations of mixing coefficients $\alpha, \beta, \gamma$. Note that the case of $\alpha= \beta = \gamma=1$ fails since no encoder feature is passed to the decoder. The `Learnable' setting outperforms all the other settings in most cases.

\begin{table*}[!h]
    \small
    \setlength{\tabcolsep}{10.0pt}
    \caption{Ablation study on the components of FMMs on NYUv2.}
    % \vspace*{-0.2cm}
    \centering
    \begin{tabular}{lcc|cccc}
    \toprule
    \multicolumn{3}{c|}{Settings} & RMSE$(\downarrow)$ & REL$(\downarrow)$ & $\delta_1(\uparrow)$ & $\tau(\uparrow)$\\
    \midrule
    \multicolumn{3}{l|}{Baseline} & 0.335 & 0.094 & 0.922 & 0.844\\
    \multicolumn{3}{l|}{+ skip-connection} & 0.331 & 0.094 & 0.925 & 0.847\\
    \multicolumn{3}{l|}{+ normalization} & 0.333 & 0.092 & \textbf{0.926} & 0.846\\
    \multicolumn{3}{l|}{+ learnable $\alpha, \beta, \gamma$} & 0.330 & 0.092 & \textbf{0.926} & 0.848\\
    % \multicolumn{3}{l|}{+ R2MC (without \bL)} & 0.326 & 0.092 & \textbf{0.926} & 0.857\\
    \multicolumn{3}{l|}{+ R2MC} & \textbf{0.325} & \textbf{0.091} & \textbf{0.926} & \textbf{0.858}\\
    \bottomrule
    \end{tabular}
    \label{tb:ablation_FMM_supp}
\end{table*}

\begin{table*}[!h]
    \small
    \setlength{\tabcolsep}{10.0pt}
    \caption{Impacts of the hyperparameters in FMMs on NYUv2.}
    % \vspace*{-0.2cm}
    \centering
    \begin{tabular}{ccc|cccc}
    \toprule
    $\alpha$ & $\beta$ & $\gamma$ & RMSE$(\downarrow)$ & REL$(\downarrow)$ & $\delta_1(\uparrow)$ & $\tau(\uparrow)$\\
    \midrule
    0 & 0 & 0 & 0.335 & 0.094 & 0.922 & 0.844\\
    0 & 0 & 1 & 0.341 & 0.094 & 0.920 & 0.841\\
    0 & 1 & 0 & 0.351 & 0.104 & 0.906 & 0.842\\
    0 & 1 & 1 & 0.333 & 0.093 & 0.921 & 0.843\\
    1 & 0 & 0 & 0.337 & 0.094 & 0.924 & 0.844\\
    1 & 0 & 1 & 0.341 & 0.095 & 0.918 & 0.841\\
    1 & 1 & 0 & 0.331 & 0.093 & 0.923 & 0.847\\
    0.5 & 0.5 & 0.5 & 0.332 & 0.093 & 0.924 & \textbf{0.849}\\
    \midrule
    \multicolumn{3}{c|}{Learnable} & \textbf{0.330} & \textbf{0.092} & \textbf{0.926} & 0.848\\
    \bottomrule
    \end{tabular}
    \label{tb:ablation_coeff_supp}
\end{table*}

\vspace*{1cm}

Table \ref{tb:coeff_results} lists the trained values of $\alpha, \beta, \gamma$ in the proposed VDE for NYUv2. There is no clear tendency, so it is difficult to find these values manually.

\begin{table*}[!h]
    \small
    \setlength{\tabcolsep}{10.0pt}
    \caption{Values of $\alpha, \beta, \gamma$ after training.}
    % \vspace*{-0.2cm}
    \centering
    \begin{tabular}{c|cc|cc|cc}
    \toprule
    % \multicolumn{7}{c}{VDE (Swin-Base)}\\
    % \midrule
    & \multicolumn{2}{c|}{FMM1} & \multicolumn{2}{c|}{FMM2} & \multicolumn{2}{c}{FMM3}\\
    % \midrule
    Coefficient & 1st & 2nd & 1st & 2nd & 1st & 2nd\\
    \midrule
    $\alpha$ & 0.4759 & 0.4866 & 0.7104 & 0.6753 & 0.5989 & 0.6802\\
    $\beta$ & 0.4474 & 0.4541 & 0.5978 & 0.6129 & 0.4530 & 0.4461\\
    $\gamma$ & 0.4773 & 0.4931 & 0.5092 & 0.4808 & 0.4512 & 0.4804\\
    \bottomrule
    % \toprule
    % \multicolumn{7}{c}{VDE (Swin-Large)}\\
    % \midrule
    % & \multicolumn{2}{c|}{FMM1} & \multicolumn{2}{c|}{FMM2} & \multicolumn{2}{c}{FMM3}\\
    % Coefficient & 1st & 2nd & 1st & 2nd & 1st & 2nd\\
    % \midrule
    % $\alpha$ & 0.4728 & 0.4907 & 0.7003 & 0.6532 & 0.5865 & 0.6994\\
    % $\beta$ & 0.4463 & 0.4720 & 0.5534 & 0.5566 & 0.4537 & 0.4432\\
    % $\gamma$ & 0.4838 & 0.4943 & 0.4967 & 0.4854 & 0.4629 & 0.4801\\
    % \bottomrule
    \end{tabular}
    \label{tb:coeff_results}
\end{table*}
\clearpage

\subsection{R2MC}
In Table \ref{tb:performance_L_SUNRGBD}, we analyze the impacts of learnable matrices $\bL$ in R2MCs on SUN RGB-D. Here, `without $\bL$' means that the value matrix $\bV_R$ obtained from $\bZ_R$ via self-attention is used instead of $\bL$. We see that the setting `with $\bL$' provides better results overall, indicating that $\bL$ plays the role of a camera-specific mapping function effectively.

\vspace*{1cm}
\begin{table*}[!h]
    \small
    \setlength{\tabcolsep}{6.7pt}
    \caption{Impacts of learnable matrices $\bL$ in R2MCs on SUN RGB-D.}
    \centering
    \begin{tabular}{l|l|ccccc}
    \toprule
    Setting & Camera & RMSE$(\downarrow)$ & REL $(\downarrow)$ & $\delta_1 (\uparrow)$ & $\tau (\uparrow)$\\
    \midrule
    \multirow{5}{*}{without $\bL$} & Kinect v1 & 0.291 & \textbf{0.129} & \textbf{0.860} & 0.791\\
    & Kinect v2 & 0.226 & 0.085 & 0.942 & 0.855\\
    & RealSense & \textbf{0.192} & \textbf{0.085} & 0.949 & 0.829\\
    & Xtion & \textbf{0.330} & 0.433 & 0.844 & 0.773\\
    \cmidrule{2-6}
    & \multicolumn{1}{|c|}{Mean} & \textbf{0.254} & 0.142 & 0.898 & 0.811 \\
    \midrule
    \multirow{5}{*}{with $\bL$} & Kinect v1 & \textbf{0.289} & 0.132 & 0.855 & \textbf{0.797}\\
    & Kinect v2 & \textbf{0.220} & \textbf{0.083} & \textbf{0.945} & \textbf{0.860}\\
    & RealSense & 0.194 & \textbf{0.085} & \textbf{0.950} & \textbf{0.835}\\
    & Xtion & 0.339 & \textbf{0.397} & \textbf{0.853} & \textbf{0.783}\\
    \cmidrule{2-6}
    & \multicolumn{1}{|c|}{Mean} & \textbf{0.254} & \textbf{0.139} & \textbf{0.900} & \textbf{0.818} \\
    \bottomrule
    \end{tabular}
    \label{tb:performance_L_SUNRGBD}
\end{table*}

\clearpage

\section{More Qualitative Results}\label{qualitative}
\begin{figure*}[h]
  \centering
  \includegraphics[width=0.95\linewidth]{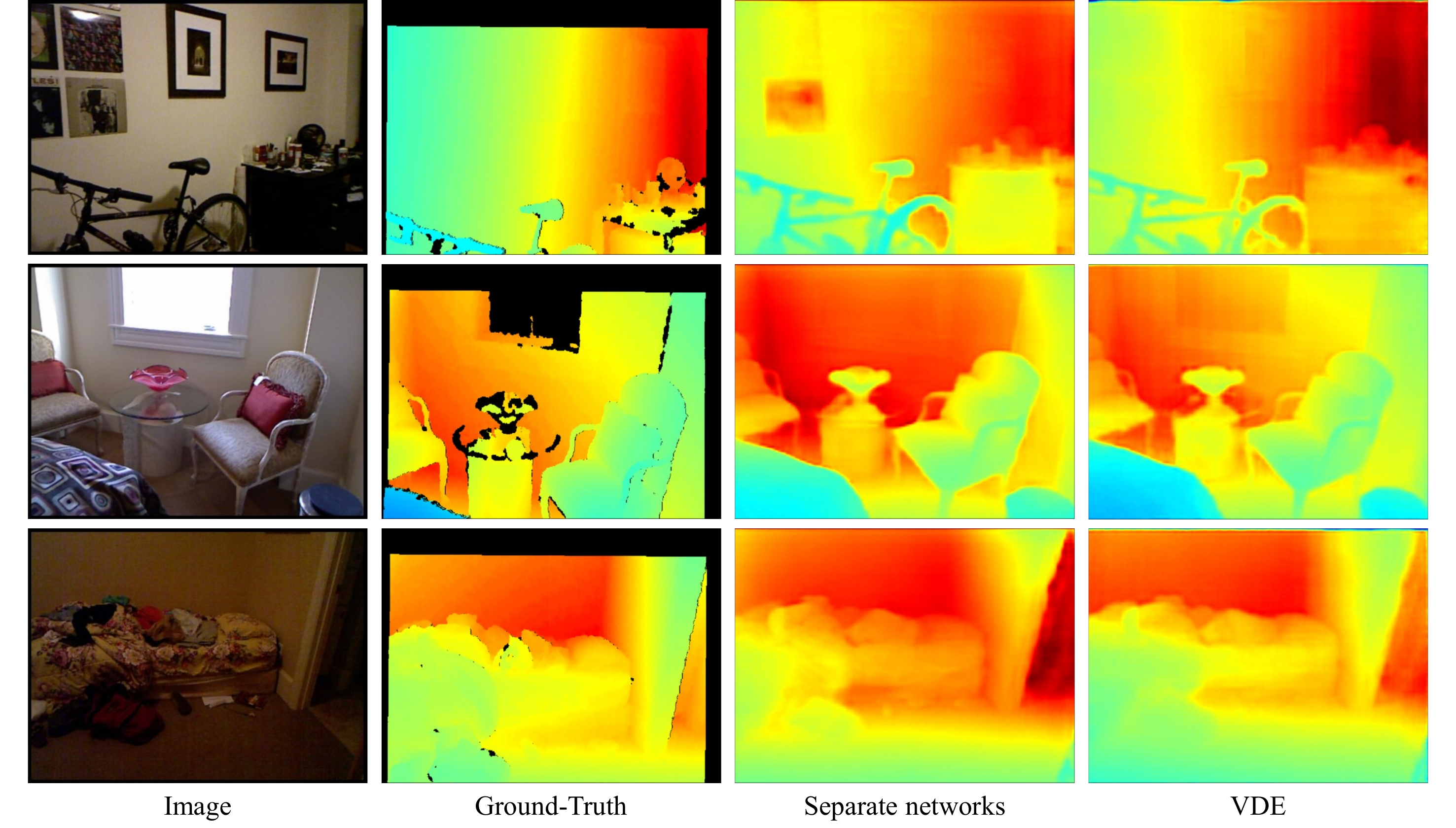}
   \caption{Versatile depth estimation reults on NYUv2.}
   \label{fig:qualitative_versatile_NYU}
\end{figure*}

\begin{figure*}[h]
  \centering
  \includegraphics[width=0.95\linewidth]{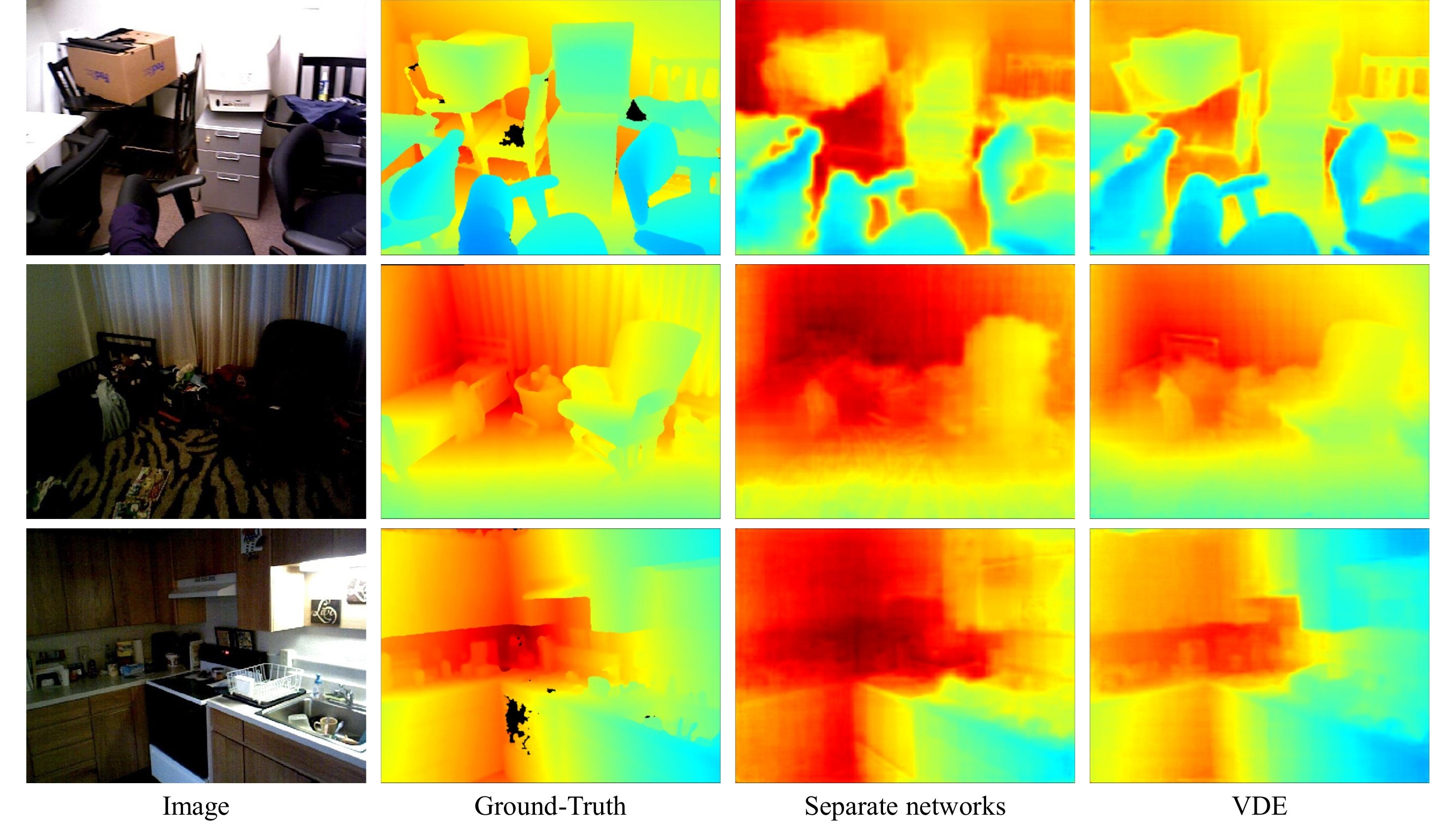}
   \caption{Versatile depth estimation reults on SUN (Xtion).}
   \label{fig:qualitative_versatile_SUN_Xtion}
\end{figure*}

\begin{figure*}[h]
  \centering
  \includegraphics[width=0.95\linewidth]{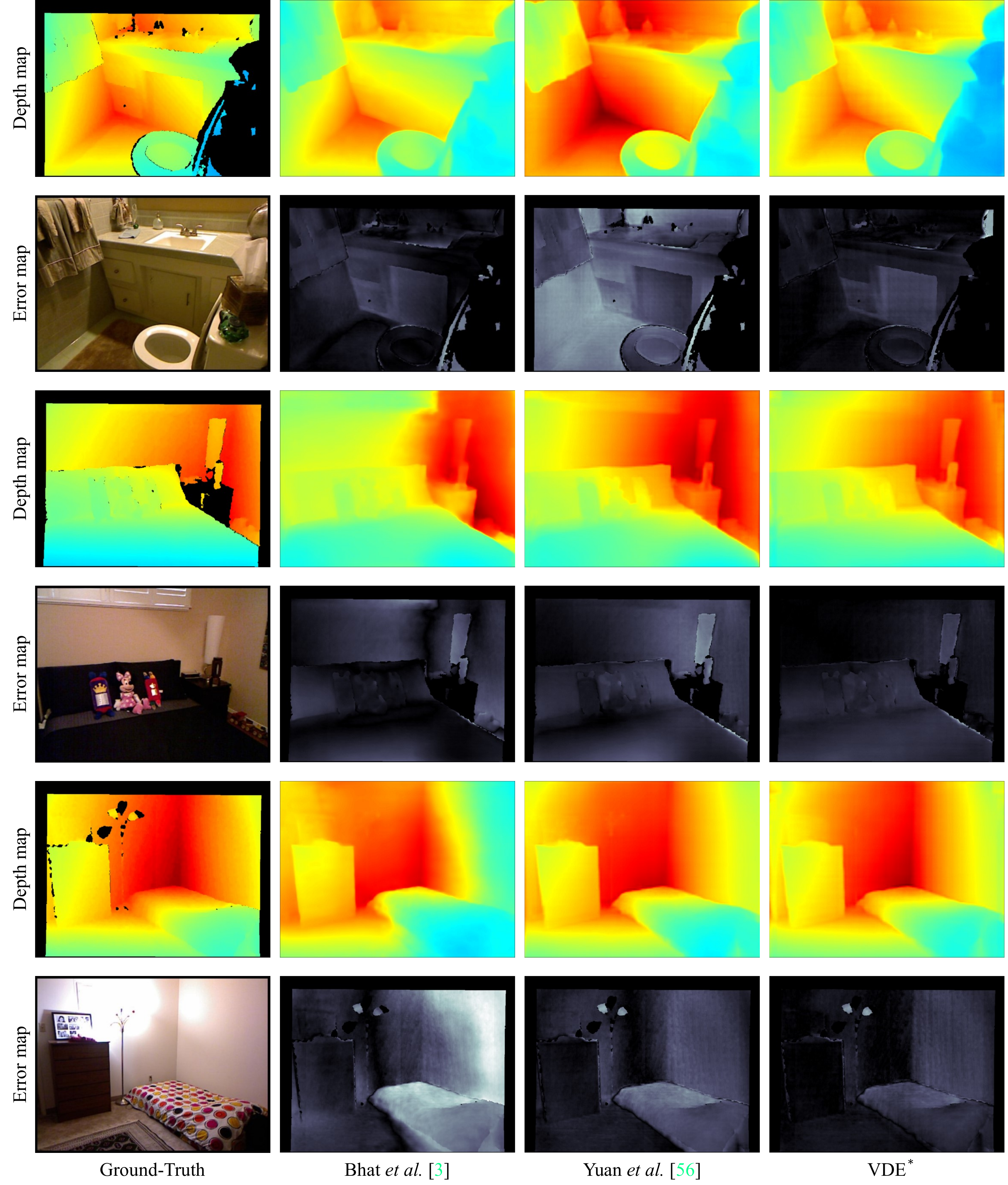}
   \caption{Qualitative comparison on NYUv2. For each depth map, the error map is also provided with a bright gray-level indicating a large error.}
   \label{fig:qualitative_nyu_supp_1st}
\end{figure*}

\begin{figure*}[h]
  \centering
  \includegraphics[width=0.95\linewidth]{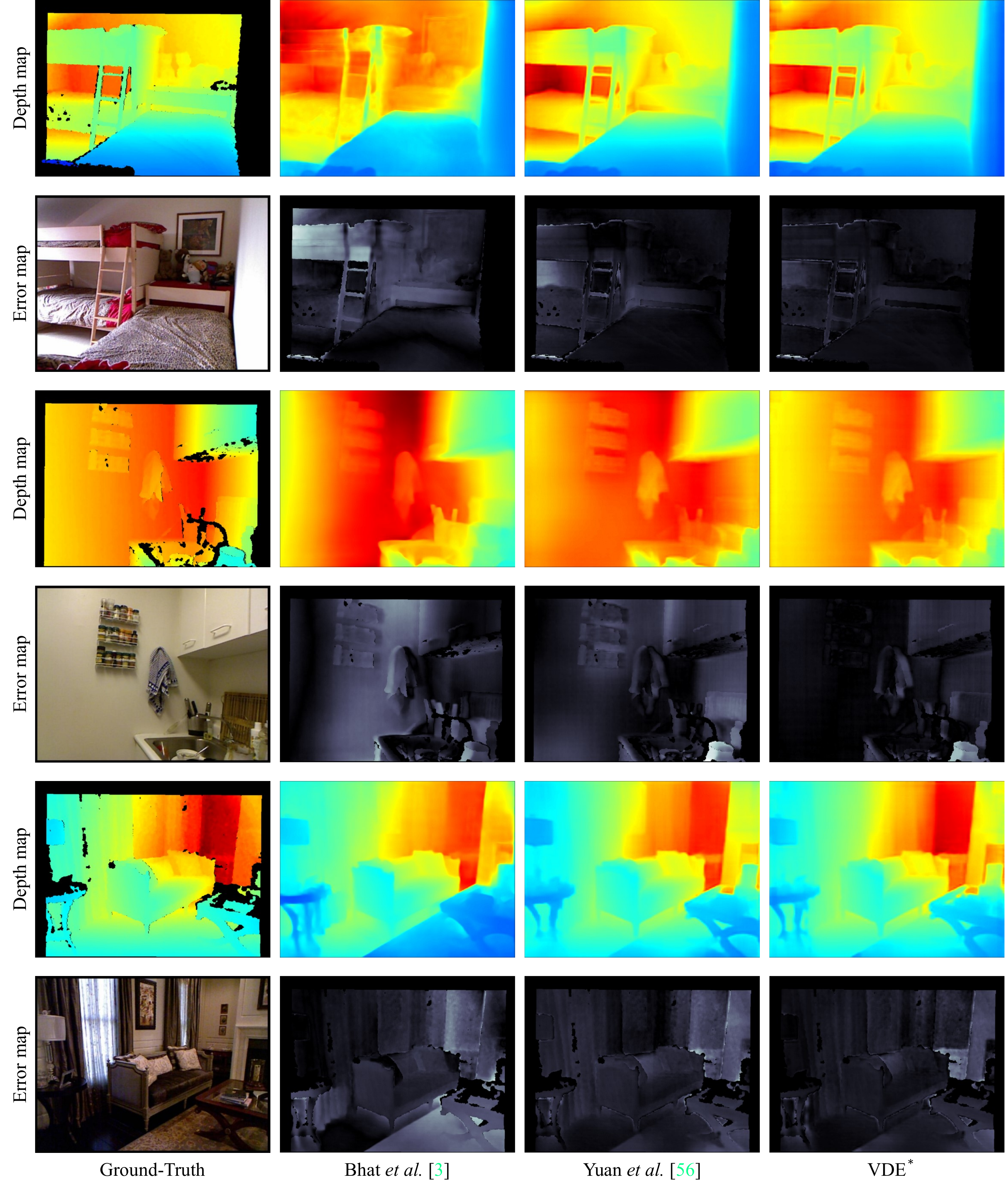}
   \caption{Qualitative comparison on NYUv2. For each depth map, the error map is also provided with a bright gray-level indicating a large error.}
   \label{fig:qualitative_nyu_supp_2nd}
\end{figure*}

\begin{figure*}[h]
  \centering
  \includegraphics[width=0.95\linewidth]{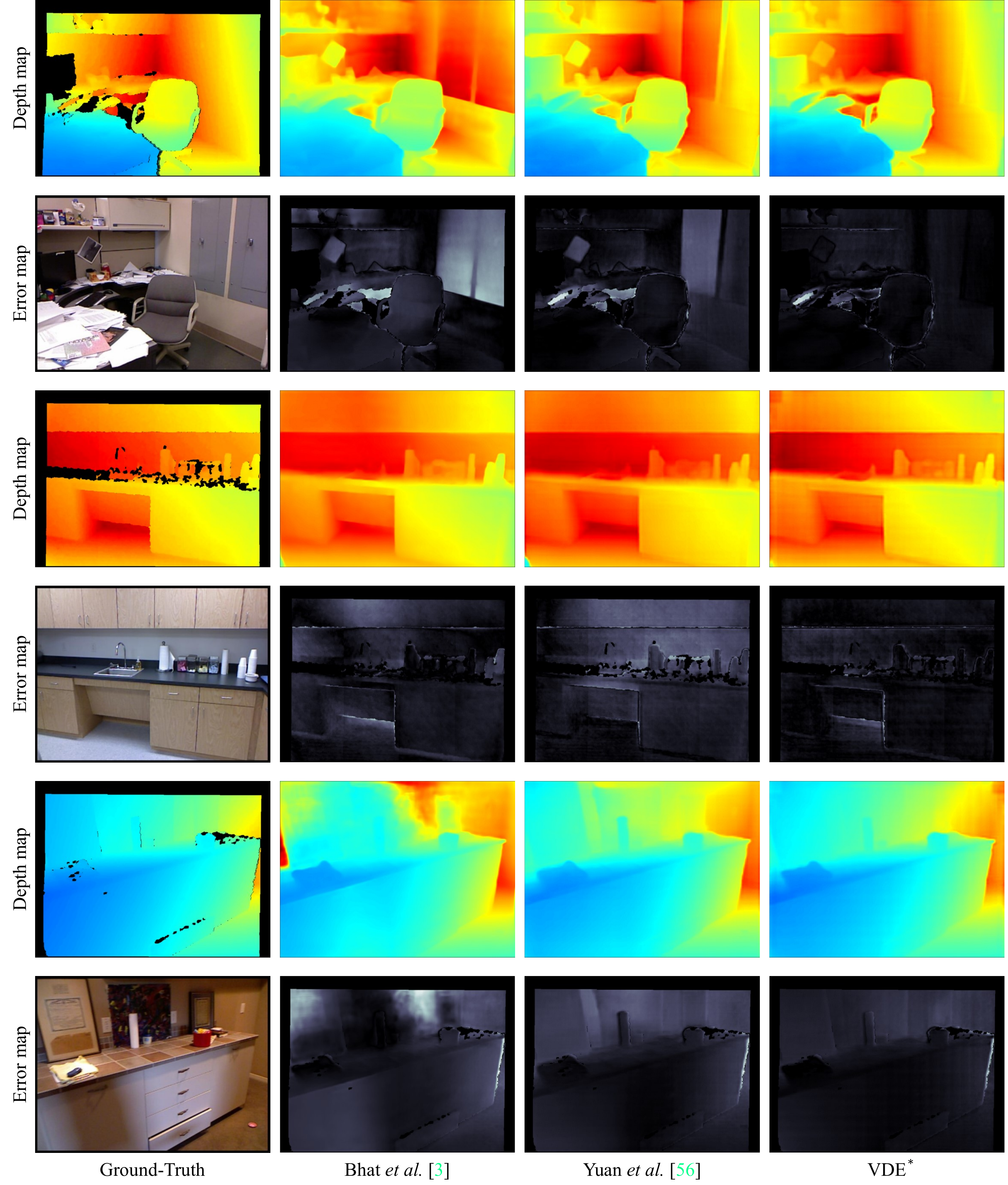}
   \caption{Qualitative comparison on NYUv2. For each depth map, the error map is also provided with a bright gray-level indicating a large error.}
   \label{fig:qualitative_nyu_supp_3rd}
\end{figure*}

Figures~\ref{fig:qualitative_versatile_NYU}$\sim$\ref{fig:qualitative_versatile_SUN_Xtion} compare versatile depth estimation results of the proposed VDE with those of `Separate networks.' Figures~\ref{fig:qualitative_nyu_supp_1st}, \ref{fig:qualitative_nyu_supp_2nd}, and \ref{fig:qualitative_nyu_supp_3rd} compare the proposed VDE with conventional algorithms \cite{bhat2021adabins, yuan2022newcrfs} qualitatively on NYUv2.

\vspace{-0.3cm}
\begin{figure*}[h]
  \centering
  \includegraphics[width=0.95\linewidth]{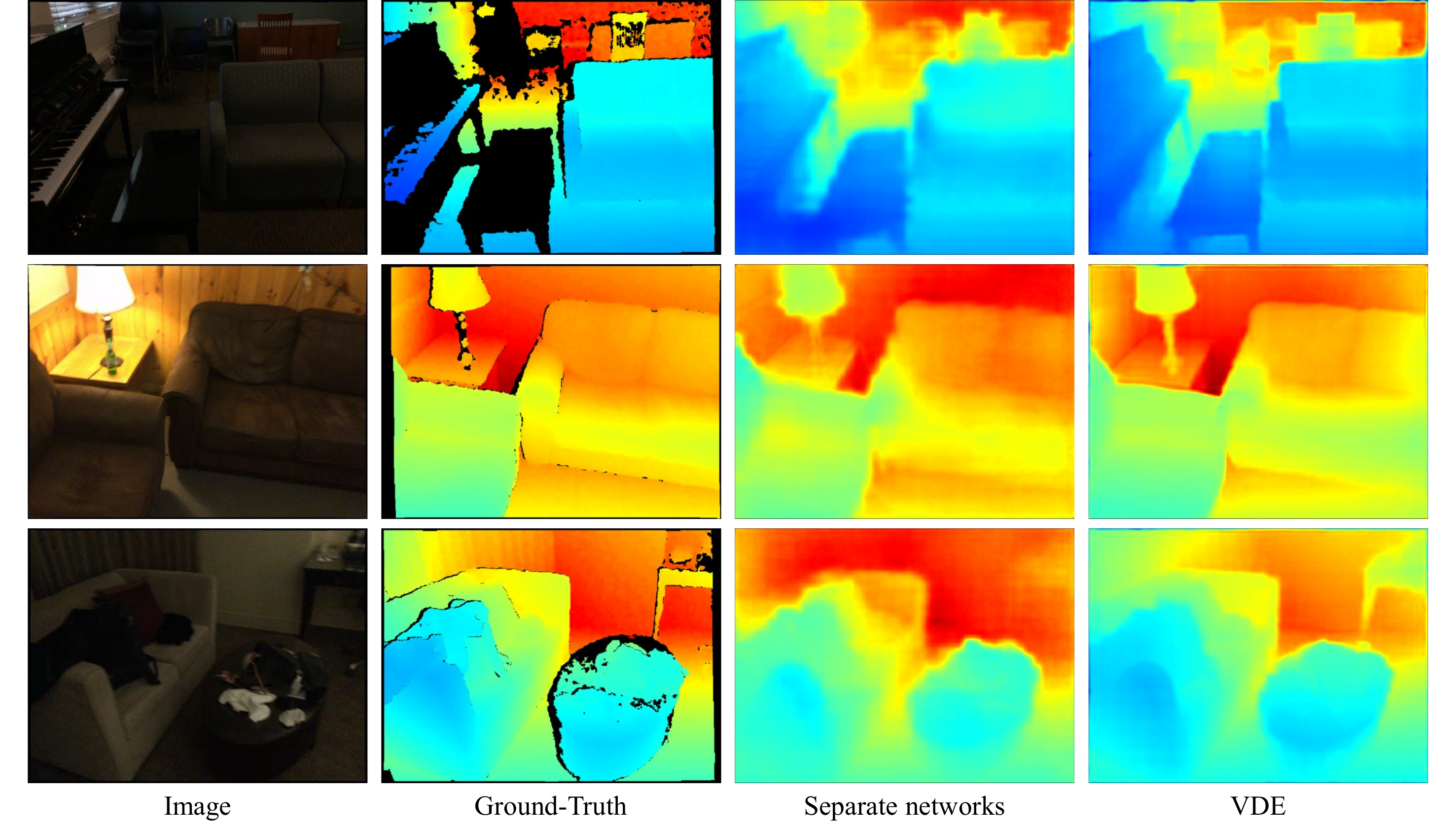}
   \caption{Versatile depth estimation reults on ScanNet.}
   \label{fig:qualitative_versatile_ScanNet}
\end{figure*}

\begin{figure*}[h]
  \centering
  \includegraphics[width=0.95\linewidth]{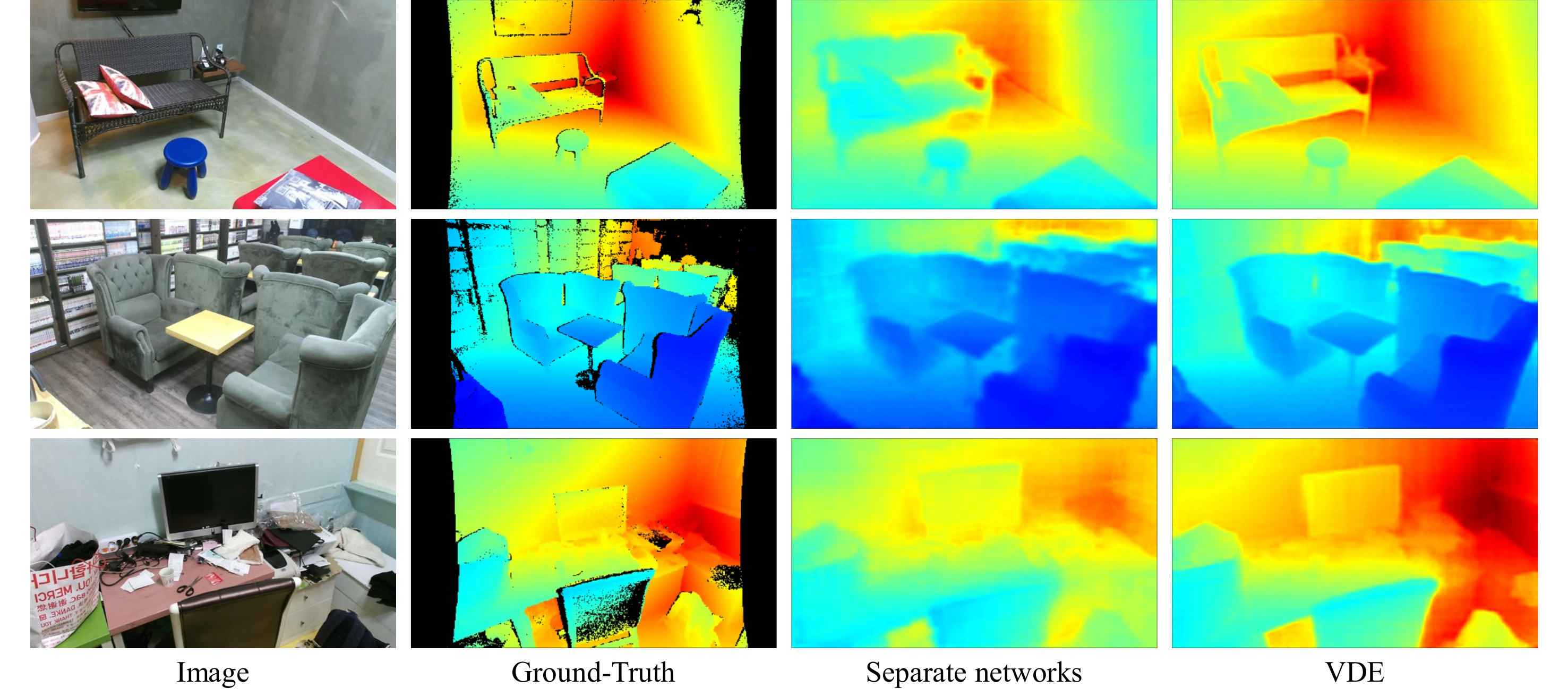}
   \caption{Versatile depth estimation reults on DIML-Indoor.}
   \label{fig:qualitative_versatile_DIML_IN}
\end{figure*}

\begin{figure*}[h]
  \centering
  \includegraphics[width=0.95\linewidth]{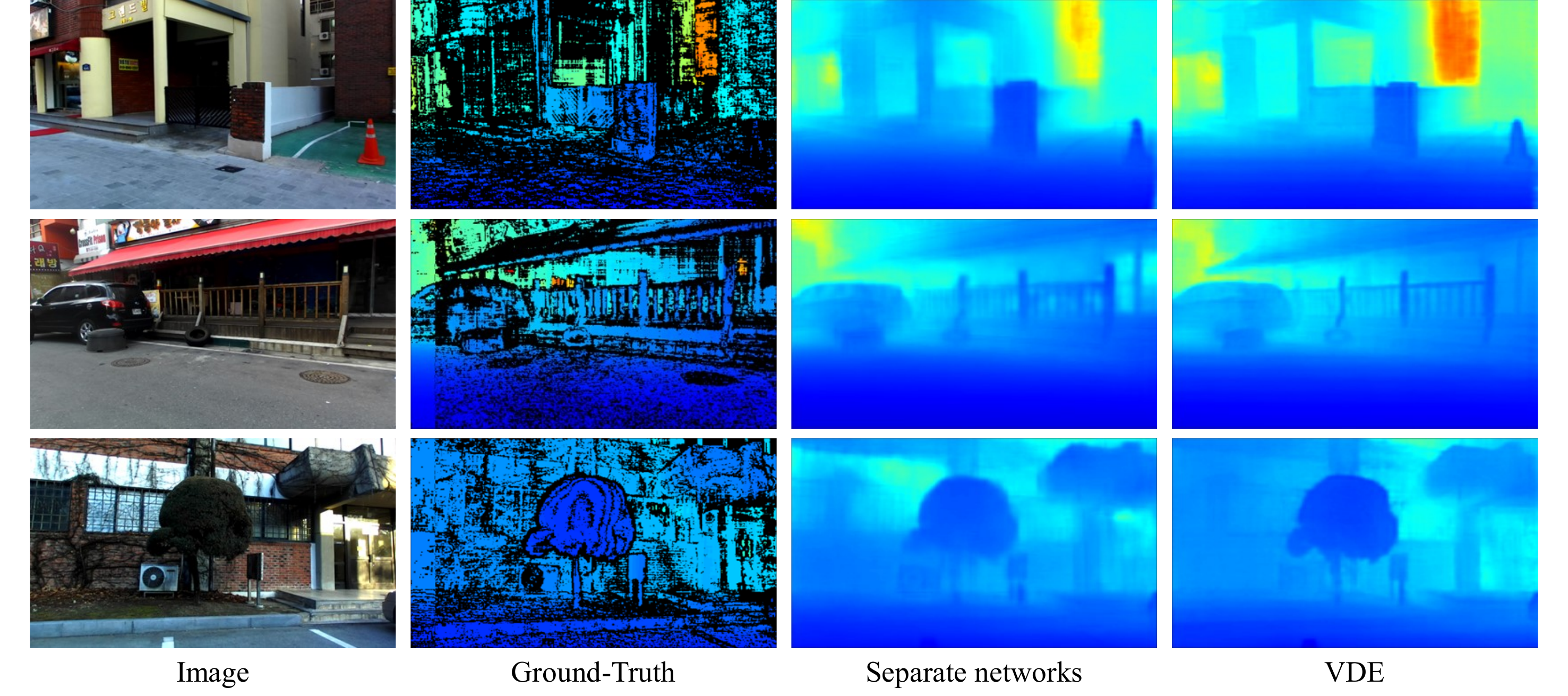}
   \caption{Versatile depth estimation reults on DIML-Outdoor.}
   \label{fig:qualitative_versatile_DIML_OUT}
\end{figure*}

\begin{figure*}[h]
  \centering
  \includegraphics[width=0.95\linewidth]{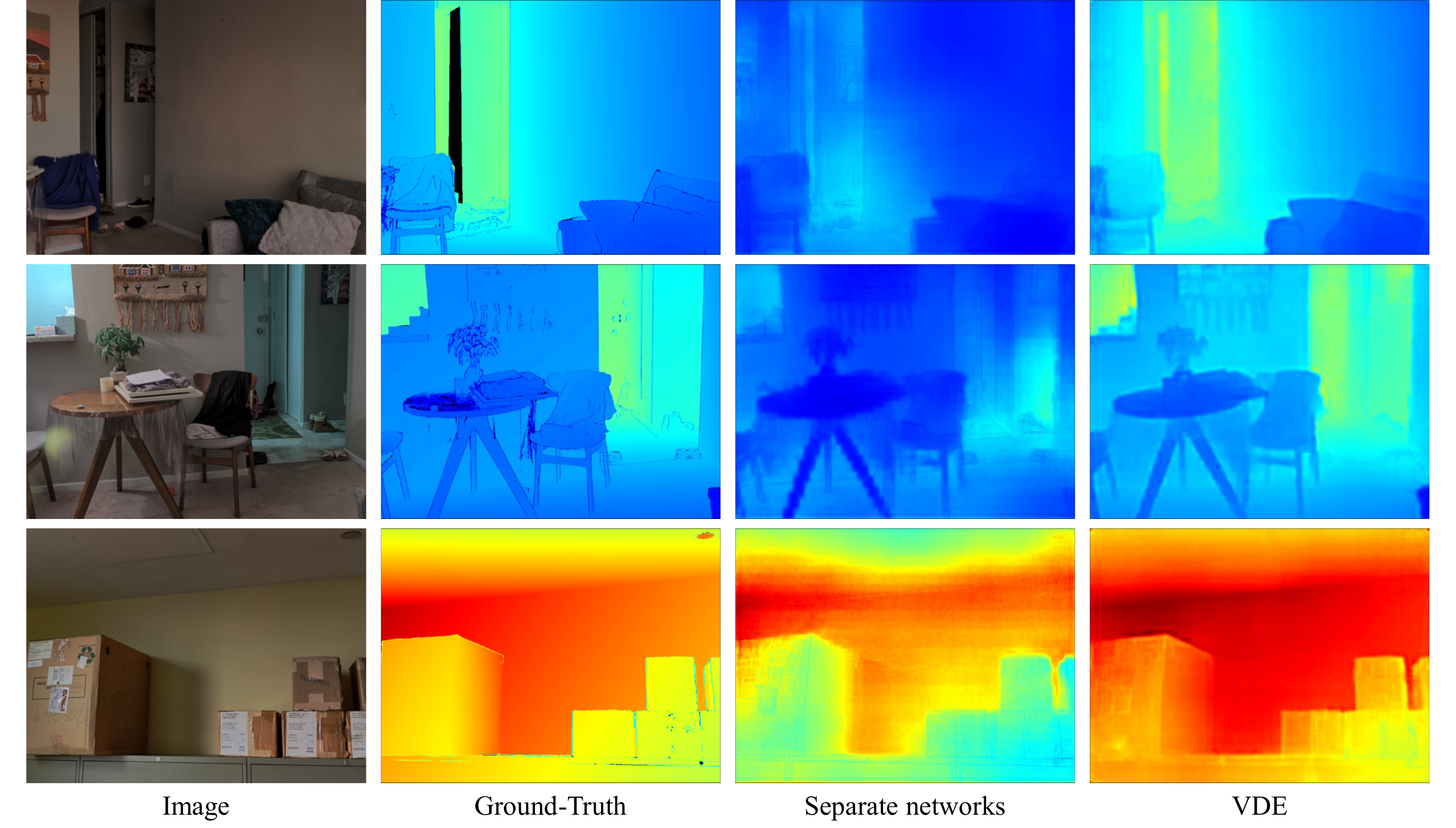}
   \caption{Versatile depth estimation reults on DIODE-Indoor.}
   \label{fig:qualitative_versatile_DIODE_IN}
\end{figure*}

\begin{figure*}[h]
  \centering
  \includegraphics[width=0.95\linewidth]{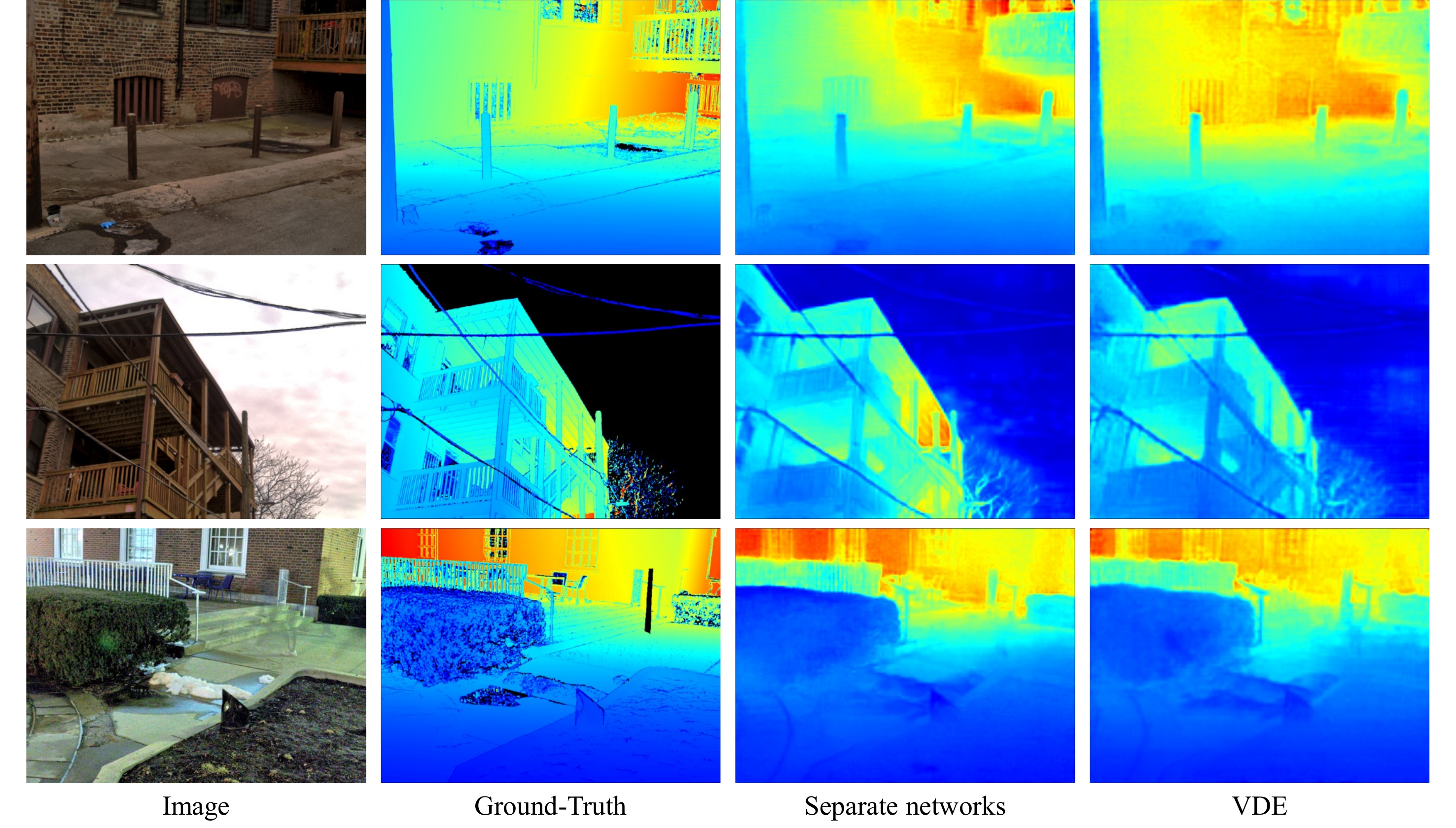}
   \caption{Versatile depth estimation reults on DIODE-Outdoor.}
   \label{fig:qualitative_versatile_DIODE_OUT}
\end{figure*}

\begin{figure*}[h]
  \centering
  \includegraphics[width=0.95\linewidth]{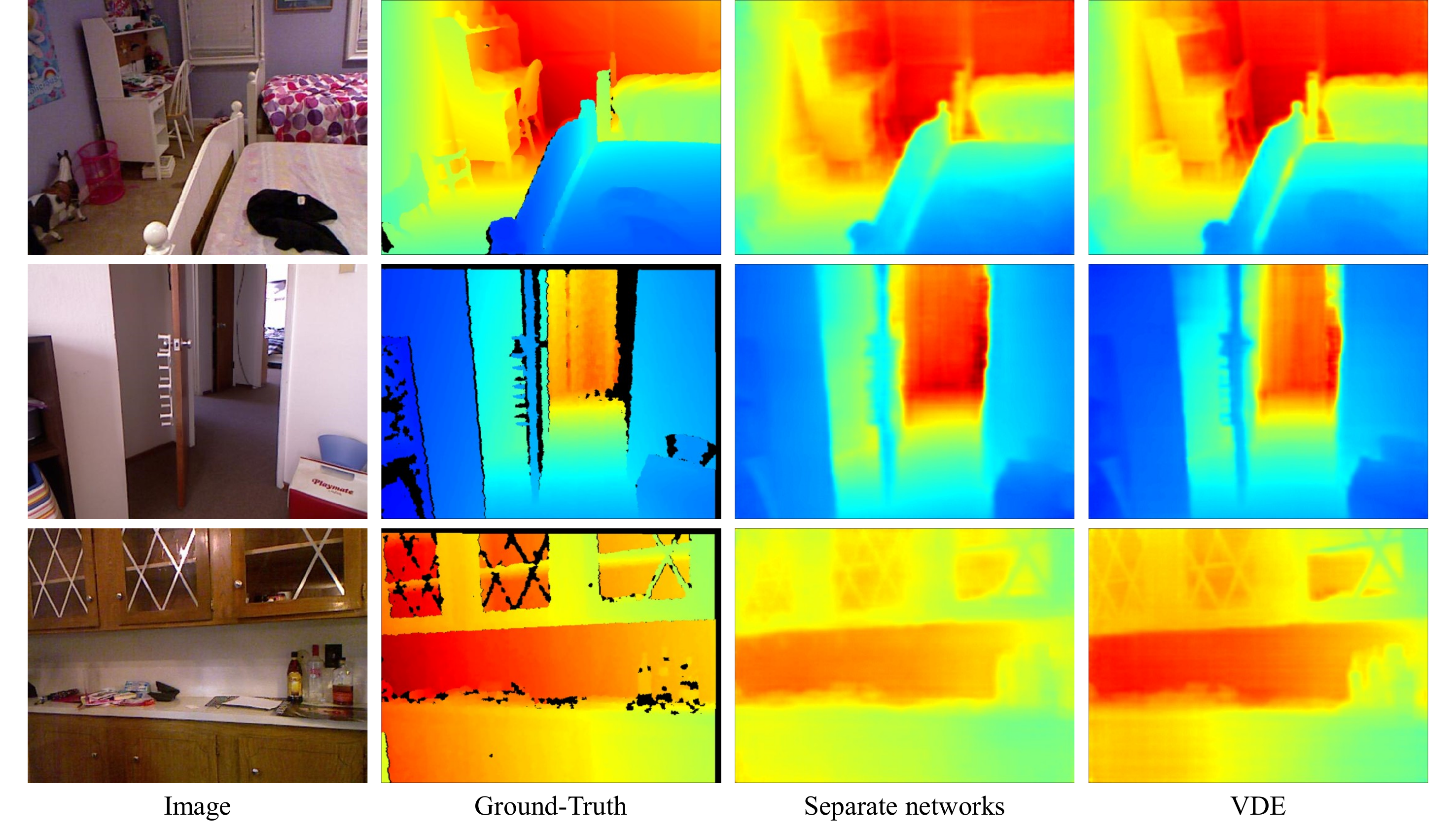}
   \caption{Versatile depth estimation reults on SUN (Kinect v1).}
   \label{fig:qualitative_versatile_SUN_KV1}
\end{figure*}

\begin{figure*}[h]
  \centering
  \includegraphics[width=0.95\linewidth]{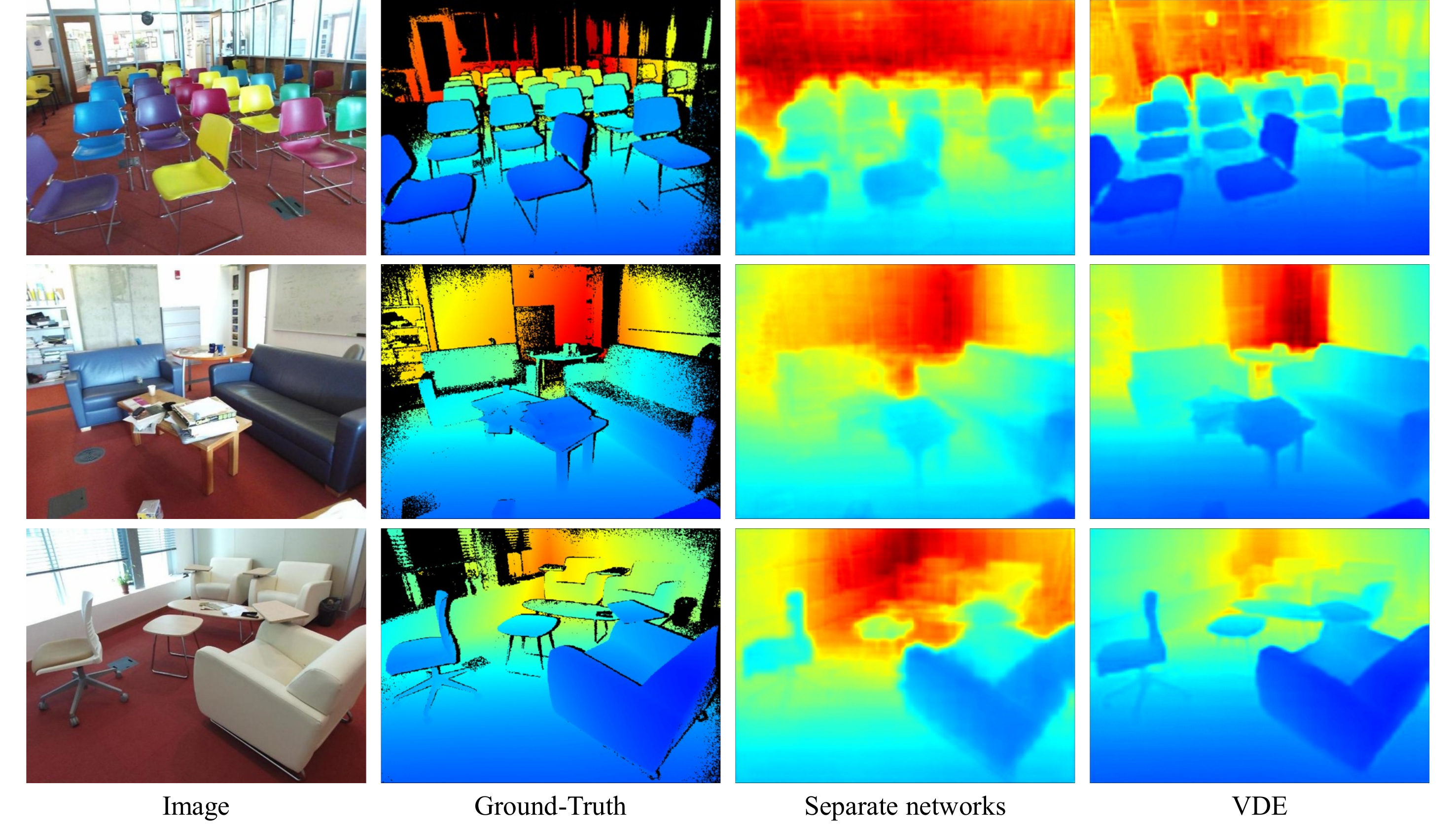}
   \caption{Versatile depth estimation reults on SUN (Kinect v2).}
   \label{fig:qualitative_versatile_SUN_KV2}
\end{figure*}

\begin{figure*}[h]
  \centering
  \includegraphics[width=0.95\linewidth]{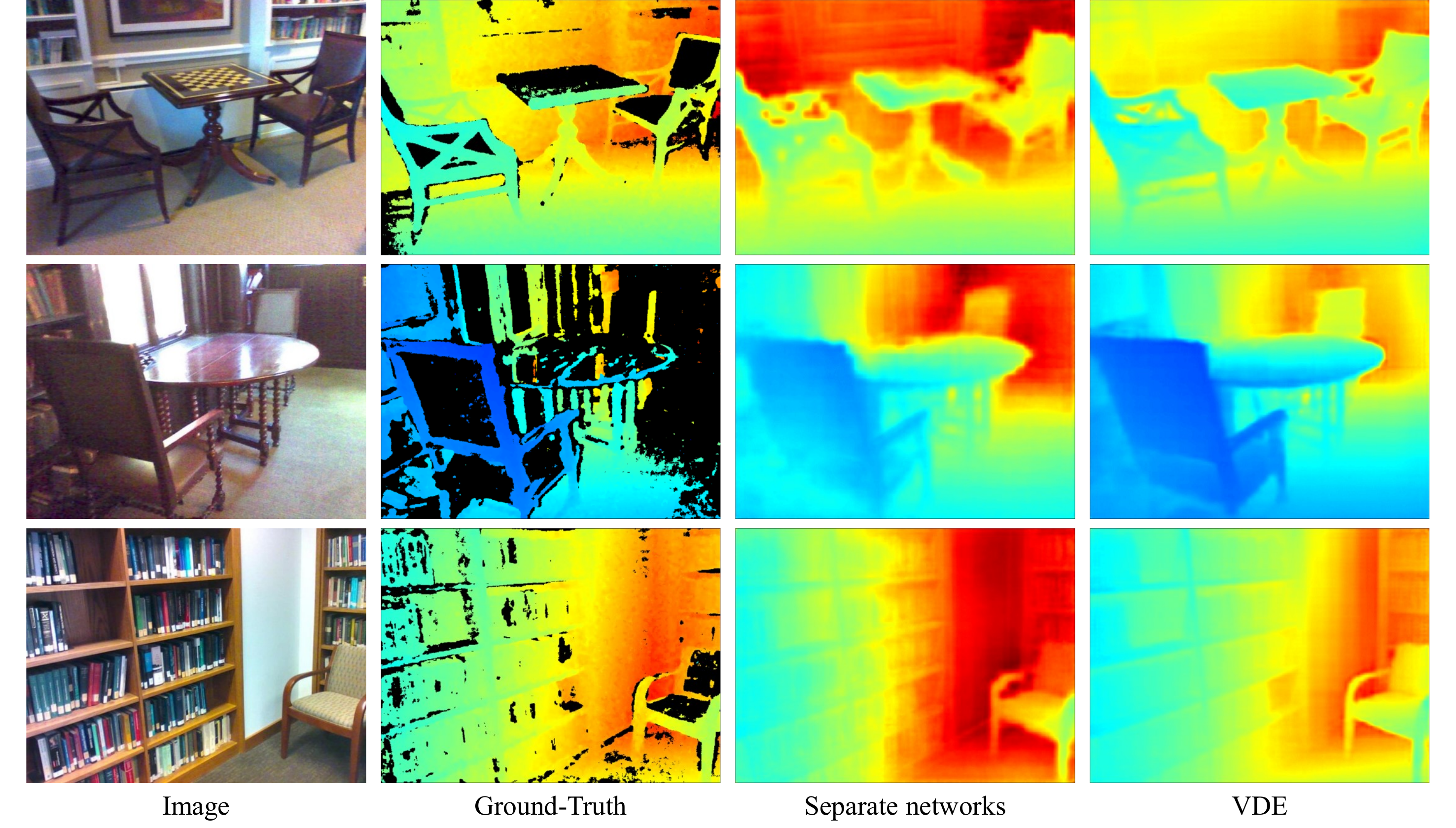}
   \caption{Versatile depth estimation reults on SUN (RealSense).}
   \label{fig:qualitative_versatile_SUN_RS}
\end{figure*}

\end{appendices}

\end{document}